\documentclass{article}
\usepackage{morse_paper,times}

\usepackage{amsmath}
\usepackage{amssymb}
\usepackage{amsthm}
\usepackage{booktabs}
\usepackage{graphicx}
\usepackage{wrapfig}
\usepackage{microtype}
\usepackage{multirow}
\usepackage{colortbl}
\usepackage{array}
\usepackage{url}
\usepackage{xspace}
\usepackage{placeins}
\usepackage{capt-of}
\usepackage[hidelinks]{hyperref}
\usepackage[most]{tcolorbox}
\usepackage{helvet}

\graphicspath{{figures_v2/}}
\definecolor{oursgray}{gray}{0.93}
\definecolor{firstc}{RGB}{212,239,213}
\definecolor{secondc}{RGB}{214,234,248}
\definecolor{thirdc}{RGB}{252,243,207}
\definecolor{kosmoBlack}{HTML}{17191D}
\definecolor{kosmoMuted}{HTML}{4B5159}
\definecolor{kosmoPanel}{HTML}{F1F4F7}
\definecolor{kosmoLink}{HTML}{315FD5}

\newcommand{\projectlink}{\href{https://kosmoresearch.github.io/MoRSe/}%
  {\textcolor{kosmoLink}{\texttt{https://kosmoresearch.github.io/MoRSe/}}}}

\renewcommand{\topfraction}{0.92}

\renewcommand{\textfraction}{0.06}

\newcommand{\method}{MoRSe\xspace}

\newcommand{\tablesetup}{%
    \scriptsize
    \setlength{\tabcolsep}{4pt}%
    \renewcommand{\arraystretch}{0.95}%
}

\newcommand{\unranked}[1]{\textcolor{black!55}{#1}}

\title{Moving Regions, Not Moving Objects: Segmenting What
Actually Moves under Camera Motion}

\hypersetup{
    pdftitle={Moving Regions, Not Moving Objects: Segmenting What Actually Moves under Camera Motion},
    pdfauthor={Wenhua Jin, Tiancheng Chen, Sheng Tang, Jiahao Li, Yupeng Liu, Weiqi Zhang, Juntong Fang, Zesong Li}
}

\makeatletter
\newif\ifkosmo@firstpage
\kosmo@firstpagetrue
\let\kosmo@iclrruler\iclrruler
\renewcommand{\iclrruler}[1]{%
  \ifkosmo@firstpage
    \makevruler[12pt][#1][1][3][0.993\textheight]%
  \else
    \kosmo@iclrruler{#1}%
  \fi}
\makeatother

\iclrfinalcopy
\begin{document}

\thispagestyle{plain}
\begingroup
\setlength{\parskip}{0pt}
\vspace*{-0.85in}
\centerline{%
\begin{minipage}{7in}
  \noindent
  \includegraphics[width=2.75cm]{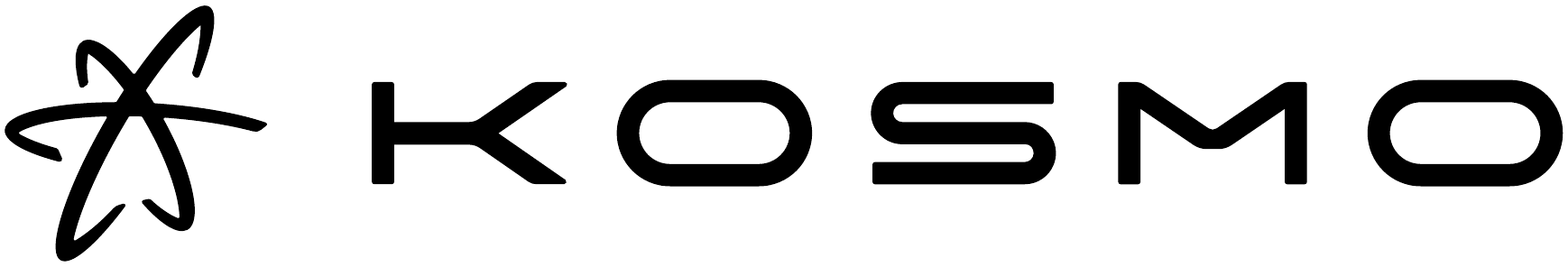}\par
  \vspace{0.15cm}
  {\color{kosmoBlack}\hrule height 0.45pt}
  \vspace{0.30cm}
  \begin{center}
    {\fontsize{20}{24}\selectfont\sffamily\bfseries\color{kosmoBlack}
      Moving Regions, Not Moving Objects: Segmenting What Actually Moves under
      Camera Motion\par}
    \vspace{0.32cm}
    {\fontsize{10.8}{13}\selectfont\sffamily\mdseries\color{kosmoBlack}
      Wenhua Jin\textsuperscript{1,2}, Tiancheng Chen\textsuperscript{1},
      Sheng Tang\textsuperscript{1}, Jiahao Li\textsuperscript{1,3},
      Yupeng Liu\textsuperscript{1,2},\\
      Weiqi Zhang\textsuperscript{4},
      Juntong Fang\textsuperscript{4}, Zesong Li\textsuperscript{1}\par}
    \vspace{0.12cm}
    {\small\fontfamily{qtm}\selectfont\color{kosmoMuted}
      \textsuperscript{1}\,Kosmo Research\quad
      \textsuperscript{2}\,Automotive Engineering Department, Jilin University\quad
      \textsuperscript{3}\,Xiamen University\quad
      \textsuperscript{4}\,School of Software, Tsinghua University\par}
    \vspace{0.26cm}
  \end{center}
  \begin{tcolorbox}[
    enhanced,
    frame hidden,
    colback=kosmoPanel,
    coltext=kosmoBlack,
    arc=8pt,
    left=0.60cm,
    right=0.60cm,
    top=0.36cm,
    bottom=0.34cm,
    boxsep=0pt,
    before skip=0pt,
    after skip=0pt,
    grow to left by=1.5pt,
    grow to right by=1.5pt
  ]
    \raggedright
    {\normalsize\sffamily\bfseries Abstract\par}
    \vspace{0.2em}
    {\fontfamily{qtm}\fontsize{9}{11.44}\selectfont
      Moving object segmentation is conventionally formulated at the sequence and object level, marking an entire instance as moving once it moves anywhere in a video.  Recent instantaneous formulations distinguish moving from stationary intervals, but still generate masks at the object level through frozen segmentation foundation models.  Because the mask is derived from an instance-level proposal, motion is typically attributed to the entire instance rather than localized to the regions that actually move.  Moreover, such instance-level predictions can produce masks even when no part of the instance is actually moving, revealing a more fundamental issue: motion is not an intrinsic property of an instance; it can only be determined by comparing at least two frames.  Building on this premise, we introduce \textbf{\method}, which jointly interprets two frames to predict a \emph{moving-region} mask, distinguishing actual scene motion from apparent motion induced by the camera.  A single RGB encoder replaces multi-stage foundation stacks, so no optical flow, point tracks, or separate mask generator is needed at inference.  We release \textbf{\method-5K}, a benchmark of 42 sequences with Static, Dynamic-Static, and Dynamic settings designed to evaluate the model's ability to distinguish and localize moving and stationary regions under different camera and scene motion conditions.  \method outperforms both specialized moving object segmentation methods and dense 4D reconstruction methods with motion-thresholded predictions, while producing the fewest false positives on sequences where nothing moves.\par}
    \vspace{0.40cm}
    {\fontfamily{qtm}\fontsize{9}{11.44}\selectfont
      \textbf{\sffamily Project page:} \projectlink\par}
  \end{tcolorbox}
  \vspace{0.26cm}
  \begin{center}
    \begin{minipage}{0.985\linewidth}
      \includegraphics[width=\linewidth]{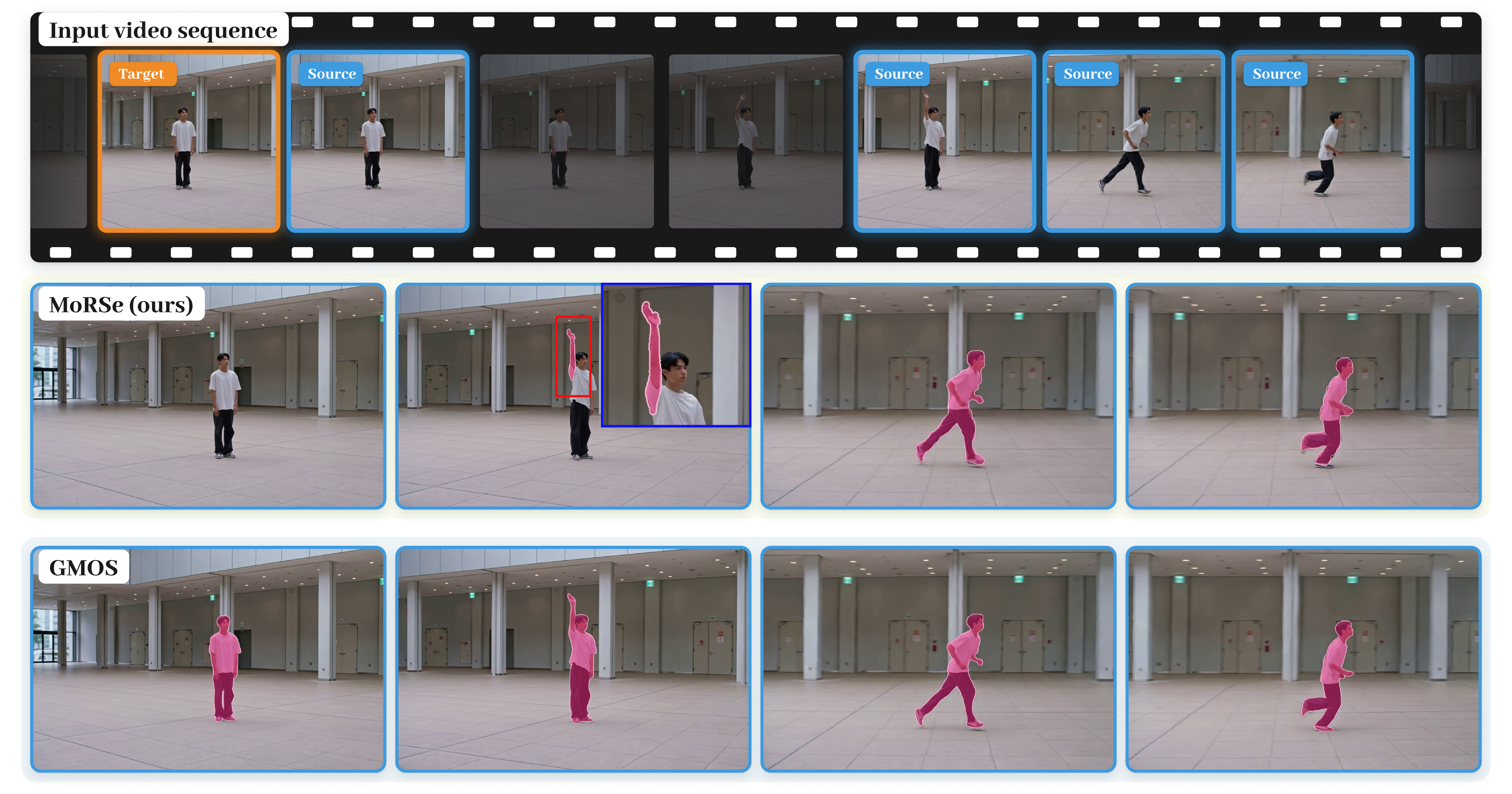}
      \makeatletter
      \long\def\@makecaption#1#2{%
        \vspace{4pt}%
        {\small\raggedright\fontfamily{qtm}\selectfont
          {\sffamily\bfseries\color{kosmoBlack}#1.}\hspace{0.5em}%
          \color{kosmoBlack}#2\par}%
      }
      \makeatother
      \captionof{figure}{\textbf{Motion is a pair relation, not an object property.}
      Motion needs two observations, so we condition on an ordered source--target
      pair.  Holding the source fixed and advancing the target through the
      sequence, \method returns an empty mask while the person stands still, then
      the moving \emph{region} alone when only an arm lifts, then the whole body
      once it walks.  GMOS marks the same person in every frame, whatever the
      target.}
      \label{fig:pair_conditioning}
    \end{minipage}
  \end{center}
\end{minipage}}
\par\vspace*{\fill}
\endgroup

\clearpage
\makeatletter\kosmo@firstpagefalse\makeatother
\fancyhead{}
\renewcommand{\headrulewidth}{0pt}

\section{Introduction}
\label{sec:introduction}

Separating independently moving regions from camera-induced image motion is a
core problem in dynamic scene understanding.  It prevents dynamic
correspondences from corrupting structure from motion and SLAM, and provides
physically grounded signals for 4D reconstruction.  Whether a region is moving
cannot be determined from its appearance alone: the same object may be
stationary for one frame pair and moving for another, while a static wall may
appear to move when the camera moves.

Existing benchmarks and models focus on moving object segmentation, treating
motion as a property of an object rather than of a frame pair.  Many video
segmentation datasets label an object throughout a
sequence once it moves at any time, so an appearance model can label a
stationary instance as moving.  GMOS takes an important step toward
instantaneous evaluation by annotating moving and stationary intervals
\citep{xie2026gmos}, but couples motion reasoning to a frozen SAM2 encoder and
an object-level motion head, which can introduce false positives for visible
but stationary objects.

We therefore define motion with respect to an ordered pair of frames.  Given a
source frame $I_s$ and a target frame $I_t$, a source pixel is moving if its
corresponding scene point changes independently of the camera between the two
frames.  The same $I_s$ should activate when the object has displaced
between the two frames, and stay suppressed when both lie in the same
stationary interval, including $I_t=I_s$.

Existing methods usually follow one of two pipelines.  Some estimate motion
with optical flow or point tracks and then use a segmentation model to complete
the mask.  Geometry explains camera motion, while the segmentation stage
recovers object extent \citep{xie2024flowsam,xie2024appearrefine,
huang2025segment,goli2025romo,he2026geomotion}.  Errors can accumulate across
correspondence, prompting, and mask propagation.  Other methods reconstruct
dense 4D motion and obtain a mask by thresholding the predicted displacement.
Feed-forward 4D models predict per-pixel motion in world coordinates that
already absorb camera motion
\citep{karhade2026any4d,luo2026forc,lin2026movies,liu2026traceanything,qian2026flow4r},
but a single displacement threshold must handle very different cases, such as a
distant vehicle and a nearby moving limb.

\textbf{\method} (\textbf{Mo}ving-\textbf{R}egion \textbf{Se}gmentation)
directly predicts a moving-region mask for each source coordinate by attending
to the target.  A UniQuery4R-initialized encoder
\citep{chen2026uniquery4r} jointly processes the selected frames alongside a
shallow convolutional branch.  The target provides motion evidence, while
multi-scale source features preserve boundary detail.  The encoder also
predicts per-view camera pose as an auxiliary task, but the predicted pose is
not used to build the mask: we do not warp features, subtract a rigid flow, or
threshold a geometric residual (Section~\ref{sec:encoder}).

Motion labels are therefore conditional on the observations rather than
attached to an object once for an entire clip.  This lets one source frame
probe three capabilities along separate axes: temporal sensitivity, by varying
the target; spatial localization, by responding only where an instance actually
moves; and robustness to egomotion, by staying silent when apparent
displacement is fully explained by the camera.  Camera motion thus becomes a
measurable negative case rather than an artifact absorbed into a
sequence-level label, motivating the benchmark and model below.

Our contributions are as follows:
\begin{itemize}
    \item We formulate motion on an ordered source--target pair and decode each
    source coordinate using target-conditioned attention, without flow, point
    tracks, or a segmentation foundation model at inference.  A single target
    gives an instantaneous region, while a target window gives a long-term
    region through a union over targets; a promptable segmenter can recover
    whole moving objects (Sections~\ref{sec:overview},
    \ref{sec:sparse_decoder}).
    \item We construct and will release \method-5K: 42 sequences and 5,431
    frames whose Static, Dynamic-Static, and Dynamic settings evaluate distinct
    model capabilities along separate axes.  The Static setting has empty ground
    truth throughout, allowing ego-motion false positives to be measured
    directly---a failure mode that sequence-level MOS benchmarks do not express
    (Section~\ref{sec:benchmark}).
    \item On \method-5K, \method performs best on moving-region segmentation
    and remains competitive on conventional sequence-level MOS benchmarks
    outside its target formulation (Section~\ref{sec:results_morse5k}).
\end{itemize}

\section{Related Work}
\label{sec:related_work}

\subsection{Flow- and Track-Based Segmentation}

Moving object segmentation (MOS) identifies objects that move in a video; in
the single-foreground setting it becomes a binary motion-segmentation problem
similar to ours.  It differs from unsupervised video object segmentation
(UVOS), which targets salient foreground objects.  The two tasks share
benchmarks such as DAVIS \citep{perazzi2016davis}, whose sequence-level labels
can reward appearance shortcuts.  GMOS adds frame-level moving/static states
and false-positive penalties \citep{xie2026gmos}, but uses a frozen geometric
encoder \citep{wang2026pi3}, frozen SAM2 features, an object-level head, and
mask propagation; its binary GMOS-S variant retains this design.  We instead
condition the mask on an ordered source--target pair.  Other MOS systems first
obtain motion proposals and then complete them with a segmenter: FlowSAM uses
flow with SAM \citep{xie2024flowsam}, Appearance-Based Refinement corrects
incomplete masks \citep{xie2024appearrefine}, and SegAnyMo prompts SAM2 from
long-term point trajectories \citep{karazija2024trajectories,huang2025segment,ravi2025sam2}.
Our model learns pair-conditioned evidence and mask prediction jointly.

\subsection{Geometric Motion Reasoning}
\label{sec:related_geometric}

Geometric methods test whether correspondences agree with a static scene under
camera motion.  RoMo fits a per-video classifier to flow-based epipolar
residuals \citep{goli2025romo}; the cue is interpretable but requires robust
fitting and test-time adaptation and is less reliable under pure rotation or
low parallax.  Easi3R derives dynamic cues from cross-view attention in a
pretrained reconstruction model \citep{chen2025easi3r}, while VGGT4D reads
attention maps from a frozen VGGT \citep{wang2025vggt} without extra training
\citep{hu2026vggt4d}.  These cues were learned for reconstruction, so a
separate statistic is needed to obtain a mask.  GeoMotion combines latent
geometry, appearance, and precomputed flow \citep{he2026geomotion}; MegaSaM
estimates motion probabilities from camera and depth \citep{li2025megasam},
MonST3R \citep{zhang2025monst3r} and DAS3R \citep{xu2024das3r} predict motion
regions with reconstruction, and MoRe uses mask-guided attention
\citep{fang2026more}.  These methods usually define masks per frame or sequence
to support reconstruction.  We make the source--target pair the basic input;
our geometry-pretrained encoder \citep{chen2026uniquery4r} predicts camera
parameters only as an auxiliary task, without using them to compensate mask
features.

\subsection{Scene Flow from 4D Reconstruction}
\label{sec:related_4d}

Feed-forward 4D models---Any4D \citep{karhade2026any4d}, 4RC
\citep{luo2026forc}, MoVieS \citep{lin2026movies}, Trace Anything
\citep{liu2026traceanything}, and Flow4R \citep{qian2026flow4r}---predict dense
world-coordinate geometry and motion.  Trace Anything represents each pixel
with a continuous trajectory field.  Their outputs can be thresholded into
masks, but that conversion is separate from the learned output.  \method uses
the same family of geometry-pretrained encoders \citep{chen2026uniquery4r} and
directly supervises a pair-conditioned mask.

\section{Method}
\label{sec:method}

\subsection{Problem Formulation and Overview}
\label{sec:overview}

Let $I_s,I_t\in\mathbb{R}^{H_0\times W_0\times 3}$ be a source and target
frame.  For a normalized source coordinate
$\mathbf{p}=(u,v)\in[0,1]^2$, the pair-conditioned label
$m_{s,t}(\mathbf{p})\in\{0,1\}$ indicates whether the scene point observed at
$I_s(\mathbf{p})$ moves independently of the camera between the two frames.
Camera-induced image displacement alone therefore has label zero.  The label
is tied to an ordered pair, so changing $t$ can change $m_{s,t}$ while leaving
$I_s$ fixed.  Our model predicts the logit
$L_{s,t}(\mathbf{p})=f_\Theta(I_s,I_t,\mathbf{p})$ and probability
$\hat m_{s,t}(\mathbf{p})=\sigma(L_{s,t}(\mathbf{p}))$ on the source grid.

Real-video pair labels inherit object-level granularity from the annotated
instances, a consequence we return to in Section~\ref{sec:limitations}.

\begin{figure}[t]
    \centering
    \includegraphics[width=\linewidth]
    {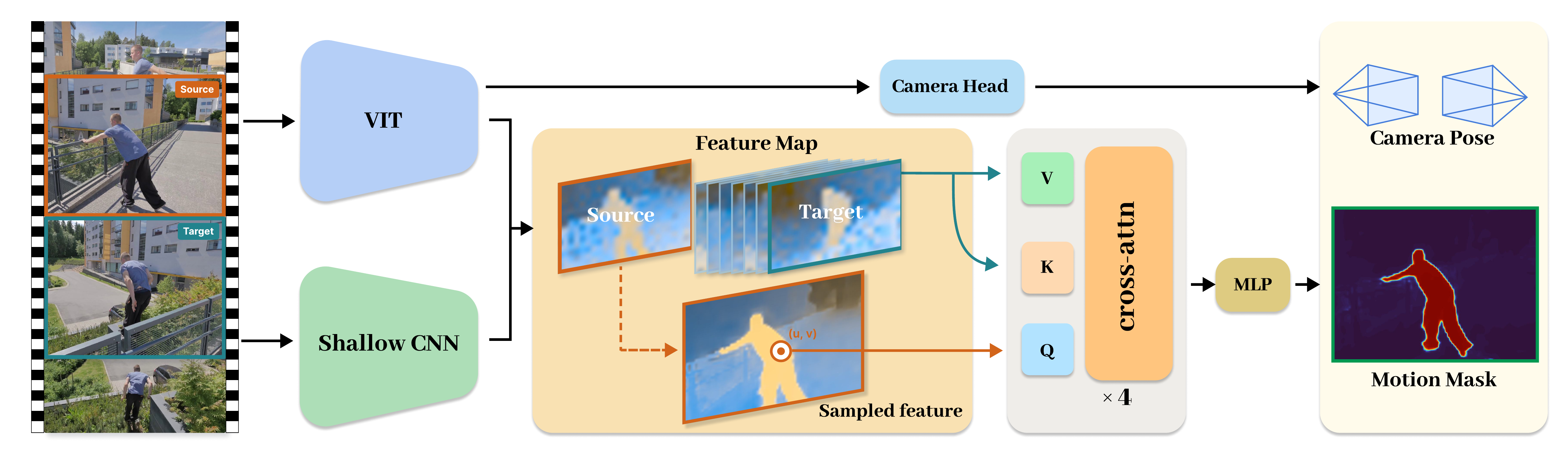}
    \caption{\textbf{\method architecture.}
    Both frames pass jointly through a UniQuery4R-initialized ViT and a
    shallow convolutional branch; the two are fused and split into source and
    target feature pyramids.  A camera head reads the ViT camera tokens and
    predicts per-view pose and field of view as auxiliary outputs that never
    enter the mask path.  Sampled source coordinates form the decoder queries;
    target features provide keys and values, and four cross-attention blocks
    refine each query before a shared MLP predicts the motion logit.}
    \label{fig:architecture}
\end{figure}

The architecture has three parts (Figure~\ref{fig:architecture}): a joint
encoder for the frame pair, sampled source coordinates that become decoder
queries, and cross-attention blocks that update each query from target
features before a final motion-logit readout.

\FloatBarrier

\paragraph{Temporal granularity at inference.}
Because the pair is the smallest unit in which motion is defined, coarser
granularities are obtained by querying different target frames without
changing the model.  For a source frame $s$ and any target set
$\mathcal{T}$,
\begin{equation}
    \label{eq:granularity}
    \hat m_{s,\mathcal{T}}(\mathbf{p})
    = \max_{t\in\mathcal{T}}\ \hat m_{s,t}(\mathbf{p}),
\end{equation}

\begin{wrapfigure}{r}{0.49\linewidth}
    \vspace{-0.6\baselineskip}
    \centering
    \includegraphics[width=\linewidth]{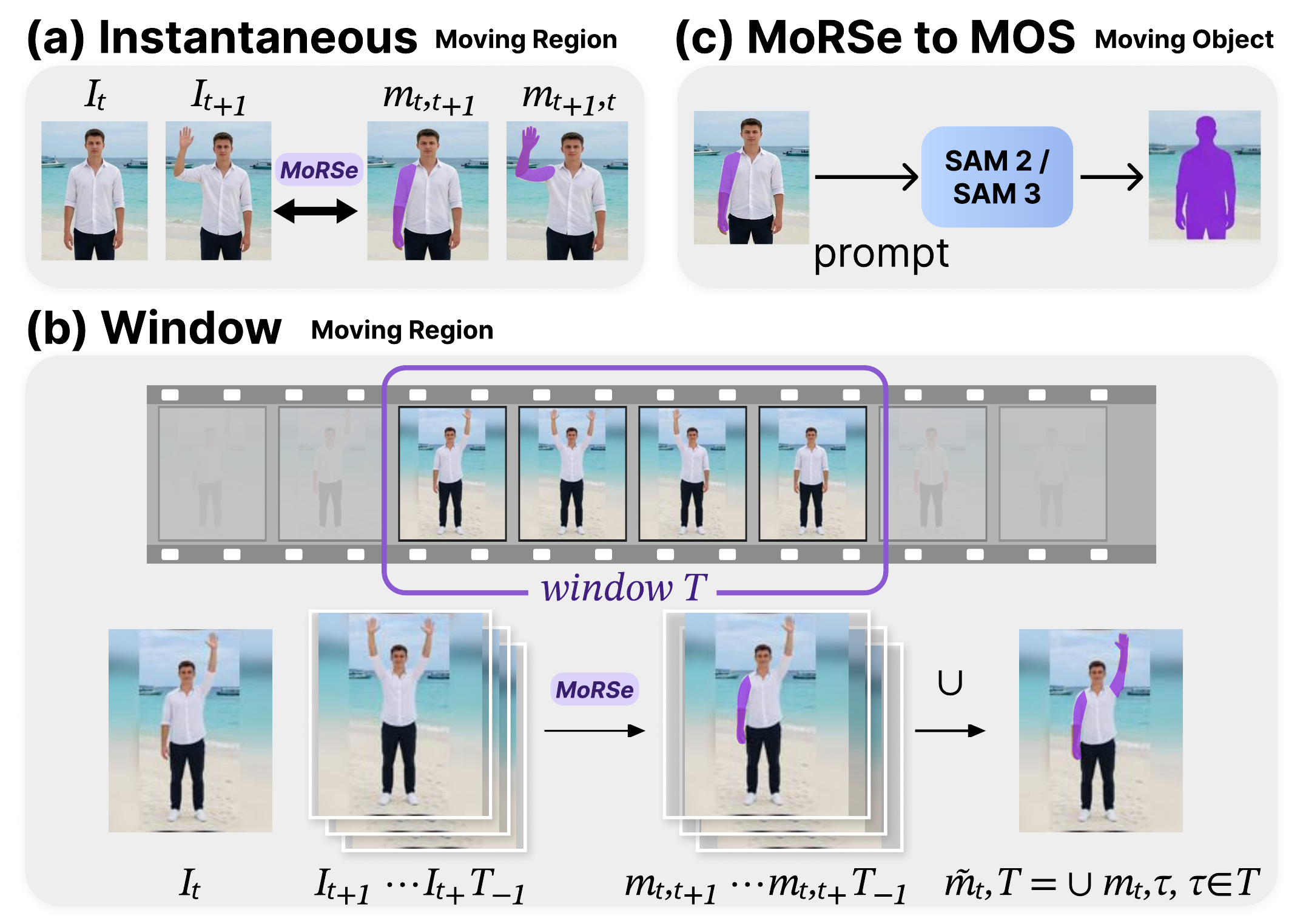}
    \caption{\textbf{Inference granularity.} Target selection sets the
    granularity without changing the weights: one pair, a target window, or a
    promptable segmenter.}
    \label{fig:inference_modes}
    \vspace{-0.8\baselineskip}
\end{wrapfigure}

This aggregation takes the largest predicted probability over the targets in
$\mathcal{T}$ (Figure~\ref{fig:inference_modes}).  After thresholding, a single
target gives the instantaneous mask, a window the moving region over that window, and
$\mathcal{T}$ spanning the video the sequence-level mask that conventional MOS
benchmarks assume (Section~\ref{sec:datasets}).  Fine-grained predictions can
therefore be aggregated, whereas sequence-level training cannot recover when
motion occurred.  Adding targets cannot remove an already selected region, but
may introduce a false positive from another pair, so target selection sets the
precision--recall trade-off without retraining.  Whole-object masks follow by
prompting a segmenter \citep{ravi2025sam2,carion2026sam3}; it does not decide
whether the region is moving.

\subsection{Joint Encoder and Camera Prediction}
\label{sec:encoder}

We initialize the visual encoder from UniQuery4R \citep{chen2026uniquery4r}.  All selected frames,
source and target alike, are passed jointly through the ViT and, in parallel,
through a shallow convolutional branch that retains high-resolution detail the
patch-tokenized ViT discards.  The two branches are fused and split into
per-view features,
$(\{F_s^{(\ell)}\}_{\ell=1}^{K},\{F_t^{(\ell)}\}_{\ell=1}^{K},c_s,c_t)
=E_\phi(I_s,I_t)$, where
$F_v^{(\ell)}\in\mathbb{R}^{H_\ell\times W_\ell\times C_\ell}$ is the
level-$\ell$ feature map, $c_v$ is a camera token, and the ordering is
fine-to-coarse so $F^{(K)}$ has the smallest spatial resolution.

A lightweight camera head reads the ViT camera tokens and predicts a per-view
pose together with a scalar field of view,
$(\widehat T_v,\widehat\phi_v)=H_{\mathrm{cam}}(c_v)$, following the
initialization's native design.  This auxiliary task uses the shared encoder,
but its predictions never enter the mask decoder; we compute no rigid flow,
epipolar residual, or compensated feature map.

\subsection{Target Memory}
\label{sec:pair_relation}

Rather than summarizing the pair into a single source-indexed map, we keep the
target features as an explicit memory that every query reads individually.  Let
$T=\operatorname{vec}(F_t^{(K)})\in\mathbb{R}^{N_t\times C}$ be the flattened
coarsest target field, projected pointwise by a layer-normalized linear map with
a GELU nonlinearity.  Each cross-attention block forms its own keys and values
$K_T,V_T$ from this memory.  The target memory is constructed once for the
pair and reused by all source queries.

Query-resolution reads matter when a small moving part and a nearby static
region fall in the same coarse feature cell.

The decoder additionally receives the source-indexed fields
$\mathcal{X}_{s}=\{F_s^{(1)},\ldots,F_s^{(K)}\}$.  At each queried coordinate,
these fields are sampled and fused into the initial query state, while the
target memory supplies the keys and values used by the cross-attention blocks.

\subsection{Sparse Query Decoder}
\label{sec:sparse_decoder}

For every source feature field $F_s^{(\ell)}$, let $\tau_\ell(\mathbf{p})$ map a
normalized image coordinate to the corresponding continuous coordinate in that
field.  Each field is projected to a common channel width, upsampled,
layer-normalized, and refined by a small convolutional stack.  Bilinear
sampling at $\tau_\ell(\mathbf{p})$ gives a per-level vector
$z_\ell(\mathbf{p})$.  The samples are merged from fine to coarse by gated
hierarchical fusion: each level mixes the running state with the incoming
sample using a per-channel sigmoid gate, followed by layer normalization and a
feed-forward update.  A final layer-normalized linear projection gives the
initial query state $q^{(0)}(\mathbf{p})\in\mathbb{R}^{C}$.
The gate lets each channel decide how much fine detail and how much coarse
context to admit; Section~\ref{sec:ablation} measures it against additive
fusion and concatenation.

The query is then refined by $B=4$ pre-normalized cross-attention blocks.  Each
block reads the target memory $K_T,V_T$ and applies a pointwise feed-forward
update at $\mathbf{p}$.  A layer-normalized linear readout maps
$q^{(B)}(\mathbf{p})$ to a feature, and the motion-mask logit is one scalar
output of a shared multi-task MLP head applied to it,
$f_\Theta(I_s,I_t,\mathbf{p})=h_{\mathrm{mask}}(q^{(B)}(\mathbf{p}))$.
Each block mixes a query with the target memory but never with another query.
Attention is computed against the fixed key/value set $K_T,V_T$, and all other
operations act on one coordinate.  A prediction is therefore independent of
which other coordinates are decoded alongside it, while a batch of $Q$
coordinates can be evaluated in parallel.  The decoder does not use
query--query attention, decoder-side spatial convolution, or coarse-to-fine
refinement, so a coordinate cannot recruit its neighbours for spatial context
(Section~\ref{sec:limitations}).

\subsection{Sampling and Objectives}
\label{sec:training}

Given a sampled clip, we supervise all directed, non-identity source--target
pairs in both temporal directions.  The supervision then follows the data
regime.  On trajectory data every query inherits the label of its tracked 3D
point: positive exactly when the point's displacement between the two frames
exceeds 7~mm in the current training recipe.  On real videos, the labels come
from the source frame's annotations---the interaction instances are positive,
the rest of the frame is negative---so the same source support keeps its label
across the targets it is paired with.  Static-scene clips contribute
all-negative supervision, which teaches the decoder to abstain when nothing
moves.

A pair is supervised on a bank of $Q=1024$ coordinates in its source frame,
composed per data regime.  Sampling changes
which coordinates receive supervision, while the prediction function remains
unchanged.  A flat mixture sampler draws clips from curated mask sources,
trajectory sources, and pseudo-static clips; only datasets with motion labels
contribute to the motion branch.  Mixture probabilities, clip lengths, and
auxiliary task weights are in Appendix~\ref{sec:impl}; curation and the split
audit are in Section~\ref{sec:training_data}.

The motion head emits one raw logit per source--target query, thresholded at
zero at inference.  The mask loss pairs a point-wise binary cross-entropy with
a soft Dice term over the valid queries,
\begin{equation}
    \mathcal{L}_{\mathrm{mask}}
    = \tfrac{1}{2}\,\overline{\operatorname{BCEWithLogits}(\ell_q,m_q)}
    + \tfrac{1}{2}\left[1
      -\frac{2\sum_q \sigma(\ell_q)m_q+1}
      {\sum_q \sigma(\ell_q)+\sum_q m_q+1}\right],
\end{equation}
where $\ell_q=f_\Theta(I_s,I_t,\mathbf{p}_q)$ and the average and sums run
over the valid queries, pooled across pairs.  Queries for points that are not
observed in the target frame receive the visibility handling specified in
Appendix~\ref{sec:impl}.  We retain the camera
objective from the initialization and optimize
$\mathcal{L}=\lambda_{\mathrm{mask}}\mathcal{L}_{\mathrm{mask}}
+\lambda_{\mathrm{cam}}\mathcal{L}_{\mathrm{cam}}$,
setting $\lambda_{\mathrm{cam}}$ to zero on datasets whose poses are not
trustworthy.  The camera loss touches only the shared encoder and camera head,
so pose accuracy stays a diagnostic rather than a proxy for mask quality.

\section{Experiments}
\label{sec:experiments}

\subsection{Training-Data Curation}
\label{sec:training_data}

Training uses 28 source entries: 19 predominantly synthetic 3D point-tracking
sources (92.3\% of sampling mass) whose motion labels threshold ground-truth
3D displacement, seven mask-source entries (6.0\%), and two static-scene
sources (1.7\%).
For mask supervision, much MOS annotation conflicts with the moving-region
definition, leaving a second mover unlabelled so that real motion becomes a
negative, or outlining an instance that is mostly still so that static pixels
become positive.  A two-stage funnel therefore retains 392 of 5,231 candidate
clips
(41,474 frames) from DAVIS 2017 \citep{ponttuset2017davis}, MoCA-Mask
\citep{cheng2022mocamask}, OVIS \citep{qi2021ovis}, YouTube-VOS 2019
\citep{xu2018youtubevos}, and HOI4D \citep{liu2022hoi4d}.
The full inventory, split provenance, and sequence-level no-leak audit are in
Appendix~\ref{sec:data_audit}.

\subsection{The \method-5K Benchmark}
\label{sec:benchmark}

\method-5K has 42 video sequences and 5,431 frames, partitioned by the
combination of camera and scene motion.  Static has a moving camera but no
moving region, so its empty ground truth measures suppression of ego-motion
false positives.  Dynamic-Static contains motion regions that intermittently
appear and disappear, while Dynamic contains camera, object, and background
motion throughout the scene.

Of these, 21 sequences (2,236 frames) are rendered in Blender \citep{blender},
with motion masks taken as pixels whose inter-frame displacement exceeds 10~mm
and then dilated.  The other 21 (3,195 frames) are real, from Mannequin
Challenge \citep{li2019mannequin}, LASIESTA \citep{cuevas2016lasiesta}, and
D3D-HOI \citep{xu2021d3dhoi}; we adopt the first two benchmarks' frame-wise
masks and annotate D3D-HOI ourselves.

\subsection{Datasets, Metrics, and Baselines}
\label{sec:datasets}

\paragraph{Metrics.}
In the Static setting every ground-truth mask is empty, so region similarity is
undefined and every predicted pixel is a false positive.  Let $\mathcal{S}$ be
the Static frames and $\hat M_i=\{\mathbf{p}:\hat m_i(\mathbf{p})>\tfrac12\}$
the moving set predicted on frame $i$ of size $H_i\times W_i$.  The pixel error
rate and the frame error rate are
\begin{equation}
    \label{eq:per_fer}
    \mathrm{PER}=\frac{\sum_{i\in\mathcal{S}}|\hat M_i|}
                       {\sum_{i\in\mathcal{S}}H_iW_i},
    \qquad
    \mathrm{FER}=\frac{1}{|\mathcal{S}|}\sum_{i\in\mathcal{S}}
    \mathbb{1}\!\left[\,|\hat M_i|>\tau_a\,\right],
    \qquad \tau_a=20\ \text{px},
\end{equation}
which separate leaking a few pixels everywhere from hallucinating a large
region occasionally.  The
Dynamic-Static and Dynamic settings are scored with the standard DAVIS region
similarity $\mathcal{J}$, contour accuracy $\mathcal{F}$, and their mean
\citep{perazzi2016davis}, pooled by frame
count (1{,}904:182), against the two ground-truth variants \method-5K provides
for the same frames; wJF denotes the pooled moving-region
$\mathcal{J}\&\mathcal{F}$, which the ablations report.  \emph{Moving region}
annotates the pixels that actually move; \emph{moving object} annotates whole
instances, an object counting as moving if any part of it does.  Rows that
emit a region or a thresholded scene flow reach the object variant through one
shared SAM~3 readout \citep{carion2026sam3}; the rest use the instance output
their pipelines already produce.  That readout expands whatever it receives,
so a false positive grows into a whole spurious instance.  Only the region
variant reads every method's native output, so we treat it as primary.

\paragraph{Scene-flow baselines.}
Trace Anything \citep{liu2026traceanything}, Any4D \citep{karhade2026any4d},
4RC \citep{luo2026forc}, and MoVieS \citep{lin2026movies} output dense motion
rather than masks, and the UniQuery4R scene-flow comparison is converted the
same way.  Every criterion is frozen when the prediction PNGs are generated and
never reselected per ground-truth variant.  Any4D and 4RC use their released
default cutoffs, and Trace Anything and VGGT4D use automatic criteria that
expose no threshold.  MoVieS is the single departure, since its official
renderer-flow value floods the frame under window-maximum aggregation, so we
calibrate $\tau=0.05$ once and freeze it.  Appendix
Table~\ref{tab:flow_provenance} lists every criterion and its provenance.

\paragraph{Other datasets and baselines.}
Section~\ref{sec:results_mos} also reports sequence-level MOS on DAVIS16-M
\citep{perazzi2016davis}, FBMS-59 \citep{ochs2014fbms}, and MoCA
\citep{lamdouar2020moca}.  Appendix~\ref{sec:protocol} states the full protocol
and baseline list.

\begin{table*}[t]
    \centering
    \caption{\textbf{Main results on \method-5K} (\%).  Metrics and the two
    ground-truth variants follow Section~\ref{sec:datasets}.  ``Scene flow'' rows
    are thresholded into masks by each model's own released or automatic
    criterion, frozen before evaluation and never re-tuned per variant.
    $\dagger$: matched-window rerun from released code.
    $\ast$: the moving-object score comes from the shared SAM~3 readout;
    unmarked rows are scored on their own instance output.
    \colorbox{firstc}{\rule{0pt}{1.4ex}best}, \colorbox{secondc}{\rule{0pt}{1.4ex}second}, \colorbox{thirdc}{\rule{0pt}{1.4ex}third} per column.}
    \label{tab:morse5k}
    \tablesetup
    \small
    \setlength{\tabcolsep}{4pt}%
    \begin{tabular*}{\textwidth}{@{\extracolsep{\fill}}llccccccccc@{}}
        \toprule
        & & \multicolumn{2}{c}{Static}
        & \multicolumn{6}{c}{Dynamic-Static $+$ Dynamic} \\
        \cmidrule(lr){3-4}\cmidrule(lr){5-10}
        & & & & \multicolumn{3}{c}{Moving region}
        & \multicolumn{3}{c}{Moving object} \\
        \cmidrule(lr){5-7}\cmidrule(lr){8-10}
        Method & Native output
        & PER$\!\downarrow$ & FER$\!\downarrow$
        & $\mathcal{J}\!\uparrow$ & $\mathcal{F}\!\uparrow$
        & $\mathcal{J}\&\mathcal{F}\!\uparrow$
        & $\mathcal{J}\!\uparrow$ & $\mathcal{F}\!\uparrow$
        & $\mathcal{J}\&\mathcal{F}\!\uparrow$ \\
        \midrule
        \multicolumn{10}{l}{\emph{Dense 4D reconstruction: predict scene flow, converted to a mask by thresholding}} \\
        Trace Anything$^\dagger$ & trajectory field & 21.01\phantom{00} & 100.00 & 32.10 & 35.90 & 34.00 & 30.18$^{\ast}$ & 38.23$^{\ast}$ & 34.21$^{\ast}$ \\
        Any4D$^\dagger$       & scene flow  & \cellcolor{thirdc}\phantom{0}0.46\phantom{00} & \phantom{0}18.70 & 27.30 & 29.80 & 28.60 & 33.80$^{\ast}$ & 38.28$^{\ast}$ & 36.04$^{\ast}$ \\
        4RC$^\dagger$         & scene flow  & \phantom{0}0.53\phantom{00} & \cellcolor{thirdc}\phantom{0}18.20 & 40.50 & 41.50 & 41.00 & 39.56$^{\ast}$ & 43.69$^{\ast}$ & 41.62$^{\ast}$ \\
        MoVieS$^\dagger$      & scene flow  & \phantom{0}0.64\phantom{00} & \phantom{0}43.80 & 39.80 & 48.50 & 44.10 & 40.42$^{\ast}$ & 49.38$^{\ast}$ & 44.90$^{\ast}$ \\
        UniQuery4R & scene flow & \cellcolor{secondc}\phantom{0}0.10\phantom{00} & \phantom{0}57.92 & 47.13 & 61.22 & 54.18 & 47.17$^{\ast}$ & 58.28$^{\ast}$ & 52.72$^{\ast}$ \\
        \midrule
        \multicolumn{10}{l}{\emph{Moving object segmentation: predict a mask directly}} \\
        FlowSAM$^\dagger$     & motion mask & 32.26\phantom{00} & 100.00 & 13.24 & 18.84 & 16.04 & 16.35 & 22.03 & 19.19 \\
        SegAnyMo$^\dagger$    & motion mask & 10.25\phantom{00} & \phantom{0}93.28 & 38.50 & 47.76 & 43.13 & 42.77 & 53.29 & 48.03 \\
        RoMo$^\dagger$        & motion mask & \phantom{0}4.22\phantom{00} & \phantom{0}27.67 & 44.85 & 51.22 & 48.03 & 46.79 & 54.76 & 50.77 \\
        Easi3R$^\dagger$      & motion mask & 12.58\phantom{00} & \phantom{0}94.05 & 33.21 & 30.39 & 31.80 & 37.34 & 33.71 & 35.52 \\
        VGGT4D$^\dagger$      & motion mask & \phantom{0}4.91\phantom{00} & \phantom{0}99.94 & 35.14 & 40.61 & 37.88 & 43.84 & 46.72 & 45.28 \\
        MoRe$^\dagger$        & motion mask & \phantom{0}9.78\phantom{00} & \phantom{0}99.89 & 50.81 & 58.24 & 54.52 & 58.76 & 64.45 & 61.60 \\
        GeoMotion$^\dagger$   & motion mask & \phantom{0}9.29\phantom{00} & \phantom{0}99.87 & 55.52 & 61.98 & 58.75 & 64.33 & 68.67 & 66.50 \\
        GMOS$^\dagger$        & motion mask & \phantom{0}1.71\phantom{00} & \phantom{0}20.37 & \cellcolor{thirdc}57.56 & \cellcolor{thirdc}63.41 & \cellcolor{thirdc}60.49 & \cellcolor{thirdc}67.41 & \cellcolor{thirdc}70.64 & \cellcolor{thirdc}69.03 \\
        GMOS-S$^\dagger$      & motion mask & \phantom{0}0.60\phantom{00} & \cellcolor{secondc}\phantom{00}6.82 & \cellcolor{secondc}66.65 & \cellcolor{secondc}72.64 & \cellcolor{secondc}69.64 & \cellcolor{firstc}\textbf{76.30} & \cellcolor{secondc}79.56 & \cellcolor{secondc}77.93 \\
        \midrule
        \textbf{\method}      & motion mask & \cellcolor{firstc}\textbf{\phantom{0}0.0009} & \cellcolor{firstc}\textbf{\phantom{00}0.75} & \cellcolor{firstc}\textbf{73.91} & \cellcolor{firstc}\textbf{82.06} & \cellcolor{firstc}\textbf{77.98} & \cellcolor{secondc}75.84$^{\ast}$ & \cellcolor{firstc}\textbf{82.79}$^{\ast}$ & \cellcolor{firstc}\textbf{79.32}$^{\ast}$ \\
        \bottomrule
    \end{tabular*}
\end{table*}

\begin{figure}[t]
    \centering
    \includegraphics[width=0.93\linewidth]
    {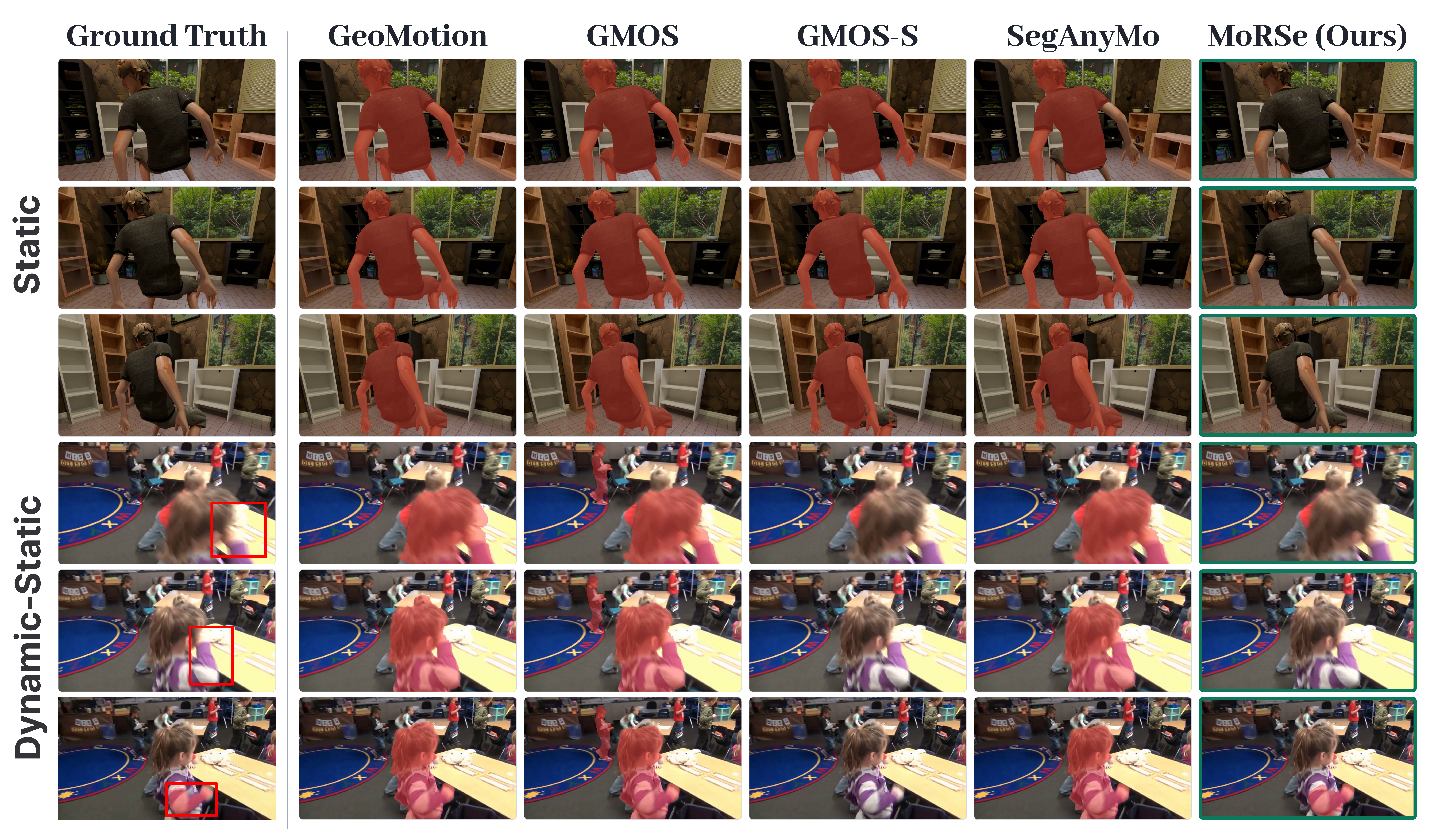}
    \caption{\textbf{Moving regions under a moving camera.}  \emph{Top
    (Static):} the camera moves while the person stands still; every baseline
    paints the whole person, while \method leaves the frame empty.
    \emph{Bottom (Dynamic-Static):} only the child's hand moves; \method marks
    that hand alone, whereas the baselines cover the whole child.}
    \label{fig:qualitative}
\end{figure}

\subsection{Moving-Region Segmentation on \method-5K}
\label{sec:results_morse5k}

\textbf{Ego-motion false positives.}  Most baselines respond to camera motion
on the Static split.  Their errors either cover many pixels or recur
across frames, whereas \method largely suppresses them (Table~\ref{tab:morse5k}).
Figure~\ref{fig:qualitative} shows the same behavior qualitatively.
\textbf{Region accuracy.}  \method gives the strongest moving-region scores,
with a particularly clear advantage on boundaries over overlap.
\textbf{Granularity.}  The advantage is smaller for moving-object evaluation,
where completing an instance from partial motion evidence favors GMOS-S on one
of the two components.  \method runs at 13.7~FPS with 9.4~GB of peak memory on
one A800-80GB at bf16.

\subsection{Window Aggregation}

\begin{wraptable}{r}{0.49\linewidth}
    \vspace{-0.8\baselineskip}
    \centering
    \tablesetup
    \small
    \caption{\textbf{Union-mask inference.} Mean precision, recall, and
    $\mathcal{J}\&\mathcal{F}$ over five benchmarks when masks from
    $|\mathcal{T}|$ target frames are unioned per source
    (Equation~\ref{eq:granularity}).}
    \label{tab:union_summary}
    \renewcommand{\arraystretch}{0.9}%
    \begin{tabular}{@{}lccc@{}}
        \toprule
        $|\mathcal{T}|$ & Precision$\uparrow$ & Recall$\uparrow$
        & $\mathcal{J}\&\mathcal{F}\uparrow$ \\
        \midrule
        2 & 86.7 & 80.4 & 72.1 \\
        4 & 86.0 & 82.2 & 72.9 \\
        8 & 84.8 & 83.3 & 73.0 \\
        \bottomrule
    \end{tabular}
    \vspace{-1.0\baselineskip}
\end{wraptable}

Equation~\ref{eq:granularity} composes pairs by a
maximum, so recall cannot fall as $\mathcal{T}$ grows.  To isolate this
inference effect, we form every target set from the same cached source--target
predictions and change only which masks are unioned.
Table~\ref{tab:union_summary} gives the first three window sizes: the added
targets recover more motion at a modest cost in precision, and the mean
$\mathcal{J}\&\mathcal{F}$ stays within a point, because false positives are
largely shared across targets rather than accumulating.  Appendix
Table~\ref{tab:union_full} gives the benchmark-wise sweep.

\subsection{Conventional MOS Benchmarks}
\label{sec:results_mos}

\begin{table}[!b]
    \centering
    \caption{\textbf{Comparison with moving object segmentation (MOS) baselines.}
    Conventional foreground--background MOS under one protocol across three
    benchmarks, with instance outputs unioned before evaluation.
    Every baseline is a controlled rerun from released code, so the numbers are
    mutually comparable but need not match the original papers.  Within each
    column, \colorbox{firstc}{\rule{0pt}{1.4ex}best},
    \colorbox{secondc}{\rule{0pt}{1.4ex}second best}, and
    \colorbox{thirdc}{\rule{0pt}{1.4ex}third best}.
    $\ddagger$: \method emits a moving region rather than instances, so its row is
    shown \unranked{in grey} for reference only and is left out of the ranking;
    the object-level entry is \method $+$ SAM~3.}
    \label{tab:main_mos}
    \tablesetup
    \resizebox{\columnwidth}{!}{%
    \begin{tabular}{lccccccccc}
        \toprule
        & \multicolumn{3}{c}{DAVIS16-M}
        & \multicolumn{3}{c}{FBMS-59}
        & \multicolumn{3}{c}{MoCA} \\
        \cmidrule(lr){2-4}\cmidrule(lr){5-7}\cmidrule(lr){8-10}
        Method
        & $\mathcal{J}$ & $\mathcal{F}$ & $\mathcal{J}\&\mathcal{F}$
        & $\mathcal{J}$ & $\mathcal{F}$ & $\mathcal{J}\&\mathcal{F}$
        & $\mathcal{J}$ & $\mathcal{F}$ & $\mathcal{J}\&\mathcal{F}$ \\
        \midrule
        OCLR$^\dagger$ \citep{xie2022oclr}            & 70.25 & 69.60 & 69.92 & 61.71 & 62.87 & 62.29 & 12.08 & 17.86 & 14.97 \\
        ABR$^\dagger$ \citep{xie2024appearrefine}     & 74.02 & 71.81 & 72.91 & 73.48 & 75.32 & 74.40 & \phantom{0}8.80 & 14.71 & 11.75 \\
        FlowSAM$^\dagger$ \citep{xie2024flowsam}      & 37.56 & 38.18 & 37.87 & 40.00 & 41.89 & 40.95 & 29.81 & 24.54 & 27.17 \\
        SegAnyMo$^\dagger$ \citep{huang2025segment}   & 75.89 & 80.82 & 78.35 & 68.80 & 76.07 & 72.44 & \cellcolor{secondc}58.81 & \cellcolor{secondc}66.10 & \cellcolor{secondc}62.46 \\
        RoMo$^\dagger$ \citep{goli2025romo}           & 70.11 & 73.87 & 71.99 & 58.86 & 61.07 & 59.96 & 11.50 & 12.39 & 11.94 \\
        Easi3R$^\dagger$ \citep{chen2025easi3r}       & 65.61 & 62.15 & 63.88 & 62.87 & 60.03 & 61.45 & 23.17 & 21.32 & 22.25 \\
        VGGT4D$^\dagger$ \citep{hu2026vggt4d}         & 56.47 & 52.01 & 54.24 & 44.31 & 44.23 & 44.27 & 33.52 & 35.57 & 34.54 \\
        MoRe$^\dagger$ \citep{fang2026more}           & 52.01 & 50.75 & 51.38 & 53.29 & 48.58 & 50.93 & 11.99 & 12.47 & 12.23 \\
        GeoMotion$^\dagger$ \citep{he2026geomotion}   & \cellcolor{thirdc}84.49 & \cellcolor{thirdc}83.77 & \cellcolor{thirdc}84.13 & \cellcolor{secondc}74.78 & \cellcolor{thirdc}76.44 & \cellcolor{secondc}75.61 & 23.69 & 27.02 & 25.36 \\
        GMOS$^\dagger$ \citep{xie2026gmos}            & \cellcolor{firstc}\textbf{87.33} & \cellcolor{firstc}\textbf{88.29} & \cellcolor{firstc}\textbf{87.81} & \cellcolor{firstc}\textbf{88.10} & \cellcolor{firstc}\textbf{91.25} & \cellcolor{firstc}\textbf{89.68} & \cellcolor{firstc}\textbf{61.49} & \cellcolor{firstc}\textbf{66.40} & \cellcolor{firstc}\textbf{63.95} \\
        \midrule
        \unranked{\textbf{\method}$^\ddagger$}        & \unranked{80.25} & \unranked{81.88} & \unranked{81.07} & \unranked{78.34} & \unranked{81.79} & \unranked{80.06} & \unranked{22.96} & \unranked{25.15} & \unranked{24.06} \\
        \textbf{\method} $+$ SAM~3                    & \cellcolor{secondc}84.78 & \cellcolor{secondc}85.83 & \cellcolor{secondc}85.31 & \cellcolor{thirdc}74.11 & \cellcolor{secondc}77.08 & \cellcolor{thirdc}75.60 & \cellcolor{thirdc}45.24 & \cellcolor{thirdc}51.47 & \cellcolor{thirdc}48.35 \\
        \bottomrule
    \end{tabular}}
\end{table}

Unioning masks over adjacent targets in both directions transfers \method to
sequence-level MOS without retraining (Equation~\ref{eq:granularity}).  These
benchmarks annotate whole instances, so parts of an object that hold still
count as misses against a region output.  The shared SAM~3 readout expands the
region to the instance, which gains 4.2 points on DAVIS16-M and roughly doubles
the MoCA score, but costs 4.5 points on FBMS-59.  GMOS, which predicts
instances natively, leads all three benchmarks.

\subsection{Ablations}
\label{sec:ablation}

\begin{table}[t]
\tablesetup
\small
\caption{\textbf{Ablations on \method-5K} (\%).  Each row changes one
component.  Entries average two seeds; the default row averages repeated runs,
unlike its single run in Table~\ref{tab:morse5k}.  All ablations are trained
on a single A800 GPU.  Per-seed spread
($\pm$ std) in Table~\ref{tab:ablation_spread}.}
\label{tab:ablation_main}
\setlength{\tabcolsep}{3pt}%
\noindent
\begin{tabular*}{0.49\linewidth}[t]{@{\extracolsep{\fill}}l r r r@{}}
    \toprule
    Variant & wJF$\uparrow$ & PER$\downarrow$ & FER$\downarrow$ \\
    & & \footnotesize($\times 10^{-4}$) & \footnotesize(\%) \\
    \midrule
    \rowcolor{oursgray}
    Default (\method) & 77.39 & 0.273 & 1.45 \\
    \midrule
    \multicolumn{4}{@{}l}{\emph{Low-level branch}} \\
    \quad ResNet-18 & 77.27 & 0.348 & 2.07 \\
    \quad ResNet-34 & 77.41 & 1.448 & 2.64 \\
    \quad ResNet-50 & 78.23 & 2.428 & 2.80 \\
    \quad removed   & 76.38 & 1.898 & 2.52 \\
    \multicolumn{4}{@{}l}{\emph{Auxiliary camera head}} \\
    \quad removed   & 76.52 & 1.657 & 5.47 \\
    \bottomrule
\end{tabular*}
\hfill
\begin{tabular*}{0.49\linewidth}[t]{@{\extracolsep{\fill}}l r r r@{}}
    \toprule
    Variant & wJF$\uparrow$ & PER$\downarrow$ & FER$\downarrow$ \\
    & & \footnotesize($\times 10^{-4}$) & \footnotesize(\%) \\
    \midrule
    \multicolumn{4}{@{}l}{\emph{Encoder initialization}} \\
    \quad DA3, official   & 74.80 & 0.250 & 0.48 \\
    \quad DINOv2 ViT-g/14 & 53.89 & 225.10\phantom{0} & 14.41 \\
    \quad DINOv3 ViT-L/16 & 48.38 & 213.19\phantom{0} & 14.47 \\
    \multicolumn{4}{@{}l}{\emph{Fusion}} \\
    \quad Ungated pooling & 76.73 & 3.117 & 4.43 \\
    \quad Concat          & 76.84 & 0.789 & 2.60 \\
    \bottomrule
\end{tabular*}
\end{table}

\textbf{Low-level branch.}  Removing the convolutional branch degrades both
region quality and static-scene suppression.  ResNet replacements trade a
small change in region quality for more false positives, so the default branch
is preferable for the target operating point.

\textbf{Auxiliary camera head.}  Removing the camera objective harms both
region accuracy and background suppression, supporting its role as a useful
regularizer for camera-induced image change.

\textbf{Encoder initialization.}  Initialization is the most consequential
choice.  DINOv2 \citep{oquab2024dinov2} and DINOv3 \citep{dinov3} perform much
worse than geometry-pretrained encoders, while our initialization improves on
Depth Anything 3 \citep{da3}.

\textbf{Fusion.}  Ungated pooling and concatenation remain close to the default
within the reported seed spread; the main gain comes from retaining the
low-level branch (Table~\ref{tab:ablation_spread}).


\section{Limitations}
\label{sec:limitations}

\textbf{Boundary quality.}  Masks from a promptable segmenter inherit its
boundaries; ours are predicted directly and are softer along hair or clothing
of similar colour.

\textbf{Uniform and repetitive surfaces.}  A blank wall or tiled floor looks
alike at matched and mismatched positions, so the decoder keys on the outer
edge and fills the interior inconsistently.

\textbf{Small-displacement motion.}  When a region moves by a very small amount
between the source and target frames, \method may fail to detect it, yielding
missed detections.

\textbf{Scarcity of real-scene region-level labels.}  Benchmarks outline whole
instances once any part moves, so region-level masks on real footage must be
drawn by hand.  Synthetic corpora parse directly into motion masks from their
per-point 3D trajectories, so the bottleneck is real-scene supervision.  Most
of our training data is rendered, and its transfer is untested at scale.

\section{Conclusion}
\label{sec:conclusion}

We introduced moving-region segmentation as a pair-conditioned task, using an
ordered source--target pair as the smallest unit in which motion is defined.
\method decodes each source coordinate by attending to the target, separating
scene motion from camera-induced change without optical flow, point tracks, or a
separate segmentation foundation model at inference.  \method-5K evaluates
this distinction under three motion settings.  Experiments show improved region
accuracy and background suppression over specialized motion-segmentation and
scene-flow baselines; real-scene supervision and small displacements remain
open.

\clearpage
\section*{AI Use Statement}

Generative AI was used as an assistive tool at several stages of this work.
For the manuscript, it assisted with literature organization, language
polishing, and editing.  For the research itself, it was used as a coding and
brainstorming aid: discussing candidate architectural choices, suggesting
refactorings during model iteration, and helping implement and debug training
and evaluation code.  All design decisions were made by the authors, and every
suggestion was reviewed, tested, and accepted or rejected on the basis of
empirical results.  Generative AI did not generate data, produce experimental
results, or determine any reported finding.  The authors verified all claims
and citations and take full responsibility for the work.

\section*{Reproducibility Statement}

We will release the code, trained weights, and the pair-index files and query
banks used for training.  We will also publicly release the \method-5K benchmark, including all 42
video sequences, the Blender scene configurations used for the synthetic split,
the frame-wise motion-region annotations, and the evaluation scripts, so that
the three camera-motion settings can be reproduced exactly
(Section~\ref{sec:benchmark}).  Training objectives and query
sampling are defined in Section~\ref{sec:training}, and every
numeric constant they depend on---bank size and positive fraction, auxiliary
camera weight, target-memory resolution, decoder depth and width, the
empty-mask area threshold, and the morphological cleanup applied to the ground
truth and to the scene-flow conversion---is tabulated in
Appendix~\ref{sec:impl}.  Dataset construction and the evaluation protocol
are detailed in Section~\ref{sec:experiments}.  All other benchmarks used are
publicly available.

\section*{Ethics Statement}

This work uses publicly available video segmentation benchmarks and raises no
issues involving human subjects, private data, or harmful insights.  No new
data were collected from human participants.

\appendix
\section*{Appendix}

\section{Training-Data Construction and Split Audit}
\label{sec:data_audit}

\paragraph{Corpus composition.}
Training draws on twenty-eight source entries in three families.  Nineteen 3D
point-tracking sources store per-frame trajectories with visibility and
validity flags, and motion labels are generated online by thresholding the
ground-truth 3D displacement at $7$\,mm; they carry $92.3\%$ of the sampling
mass.  Seven mask sources store a per-frame binary moving-object mask that
seeds the query bank ($6.0\%$), and two static-scene sources carry depth,
point maps, and camera poses with neither masks nor tracks ($1.7\%$).
Table~\ref{tab:corpus} lists them.

\begin{table}[h]
    \centering
    \caption{\textbf{Training corpus inventory.}  Family headers carry the
    normalized sampling-weight share.  LASIESTA and MannequinChallenge are
    listed for evaluation provenance and excluded from training.}
    \label{tab:corpus}
    \tablesetup
    \begin{tabular*}{\columnwidth}{@{\extracolsep{\fill}}lrr@{\hspace{2.5em}}lrr@{}}
        \toprule
        Source & Seq. & Frames & Source & Seq. & Frames \\
        \midrule
        \multicolumn{6}{@{}l}{\emph{3D point tracking} --- 19 entries, 137,103 sequences, 19,260,636 frames; 92.3\% of training samples} \\
        PointOdyssey & 111 & 207,968 & Kubric4D & 808 & 48,480 \\
        CoTracker3-Kubric & 5,827 & 699,240 & Flock4D & 950 & 434,003 \\
        TAPIP3D-Kubric & 10,950 & 262,800 & DynamicReplica & 8 & 2,400 \\
        SynthVerse & 26,187 & 4,934,809 & Stereo4D & 81,962 & 12,128,259 \\
        Syn4D & 565 & 168,787 & Waymo & 5,497 & 131,928 \\
        Internal synthetic & 3,094 & 214,506 & VKITTI2 & 1,144 & 27,456 \\
        \midrule
        \multicolumn{6}{@{}l}{\emph{Moving-object masks} --- 5 training entries plus 2 evaluation-only entries; 94,199 inventory frames} \\
        DAVIS 2017 & 26 & 1,735 & HOI4D & 106 & 31,800 \\
        MoCA-Mask & 37 & 1,378 & LASIESTA & 46 & 17,625 \\
        YouTube-VOS 2019 & 184 & 4,647 & MannequinChallenge & 585 & 35,100 \\
        OVIS & 39 & 1,914 & & & \\
        \midrule
        \multicolumn{6}{@{}l}{\emph{Static scene} --- 2 entries, 9,924 sequences, 3,200,269 frames; 1.7\% of training samples} \\
        DL3DV-10K & 9,156 & 3,125,750 & Hypersim & 768 & 74,519 \\
        \bottomrule
    \end{tabular*}
\end{table}

\paragraph{Corpus provenance.}
The named point-tracking sources trace to PointOdyssey
\citep{zheng2023pointodyssey}, Kubric \citep{greff2022kubric}, CoTracker3
\citep{karaev2025cotracker3}, TAPIP3D \citep{tapip3d}, Dynamic Replica
\citep{karaev2023dynamicstereo}, SynthVerse \citep{zhao2026synthverse},
Stereo4D \citep{jin2025stereo4d}, Syn4D \citep{jiang2026syn4d}, the Waymo Open
Dataset \citep{sun2020waymo}, Virtual KITTI 2 \citep{cabon2020vkitti2}, and
the Flock4D dataset release \citep{flock4d}.  The latter is cited as a dataset
repository because no accompanying paper is currently available.
The static sources are DL3DV-10K \citep{ling2024dl3dv} and Hypersim
\citep{roberts2021hypersim}.  Names ending in ``-Kubric'' identify the
corresponding Kubric-based release; internal synthetic data is ours.

\paragraph{Mask-source curation.}
Five mask sources are curated from public video.  DAVIS 2017, MoCA-Mask, OVIS,
YouTube-VOS 2019, and HOI4D all use their official training partitions; from
the HOI4D training split we draw a pool of 1,022 clips.  A two-stage funnel
keeps 392 of 5,231 candidate clips ($43.3\%$ of DAVIS 2017, $52.1\%$ of
MoCA-Mask, $6.4\%$ of OVIS, $5.3\%$ of YouTube-VOS 2019, $10.4\%$ of the HOI4D
pool), which is where the 41,474 frames those five contribute come from.  Both stages target
the same mismatch between object-level annotation and the moving-region
definition.  One failure mode is an unlabelled second mover, which places real
motion among the negatives; the other is an outlined instance that is mostly
still, which places static pixels among the positives.  The first stage uses
long-baseline geometric motion consistency for DAVIS 2017 and MoCA, with
$\mathrm{IOU}_{\mathrm{PRED}}=0.85$, $\mathrm{CD}_{\mathrm{PRED}}=0.05$,
minimum presence fraction $0.7$, and windows $W=12/4$; OVIS and YouTube-VOS
2019 use interval-level surface-motion consistency based on IoU, surface-feature
motion, and camera-compensated centroid displacement.  The second stage is
sequence-level manual review.  Retained masks are used as moving-object support masks: annotated
instance pixels are positive on the source frame and other pixels are negative.
They are not treated as fine-grained annotations that identify independently
moving subregions.

\paragraph{No-leak audit.}
Every retained sequence is checked against the official training split and an
explicit no-leak manifest, at the sequence level and before frame or pair
sampling.  The retained DAVIS 2017, MoCA-Mask, OVIS, YouTube-VOS 2019, and
HOI4D sequences all belong to their official training partitions and share no
sequence with the corresponding evaluation sets; HOI4D is additionally never
evaluated on in this paper.  LASIESTA and
MannequinChallenge also supply the real half of \method-5K, so their training
sequences are intersected against the evaluation manifest and every implicated
clip is removed, including LASIESTA \texttt{I\_CA\_01} and \texttt{I\_MB\_01}
and the MoCA clips \texttt{crab\_2}, \texttt{grasshopper\_1}, \texttt{plaice},
\texttt{seal}, and \texttt{wolf}.  The released manifest records the source
dataset, split, sequence identifier, filtering outcome, and no-leak status for
every candidate.

\section{Computational Structure}
\label{sec:complexity}

Let $N_t$ be the number of target memory tokens, $Q$ the number of decoded
queries, and $B$ the number of decoder blocks.  Building the target memory
costs $\mathcal{O}(N_tC)$ once per frame pair, and decoding costs
$\mathcal{O}(BQN_tC)$.  Because each query attends to the target
independently, this term scales with the number of decoded coordinates, so
concentrating supervision on $Q$ informative locations is what makes
interaction at query resolution affordable to train.

These expressions cover the decoder, not the entire system.  The joint encoder
and the once-per-pair target memory are paid regardless of how many
coordinates are decoded and may dominate runtime for a large initialization.
Appendix~\ref{sec:efficiency_robustness} measures both terms separately and
sweeps $Q$, $N_t$, and the window length against wall-clock cost.

\section{Implementation Details}
\label{sec:impl}

Table~\ref{tab:hparams} lists every constant referenced in the main text in one
place: the query-sampling mixture weights, the loss balance, the auxiliary
camera weight, the target-memory resolution, the decoder width and depth, the
empty-mask area threshold $\tau_a$, and the morphological cleanup used both to
build the synthetic ground truth (Section~\ref{sec:benchmark}) and to
post-process the thresholded scene-flow baselines
(Section~\ref{sec:datasets}).  The two cleanups share a $7\!\times\!7$
elliptical structuring element but not their operators.  Ground-truth masks are
dilated ten times and clipped to the annotated instance silhouette, with no
component filter, so that a thin moving part is never erased.  Thresholded
scene-flow predictions instead receive one opening followed by one closing, and
components below 300~px are dropped, because the per-pixel magnitude otherwise
leaves isolated speckle that no baseline intends as a prediction.  That filter
discards small false and true positives alike, so on the static split---where
any prediction is an error---it works in the baselines' favour.

\paragraph{Training recipe.}
The motion-head experiments fine-tune UniQuery4R-initialized weights; the
motion-mask output head is randomly initialized and loaded non-strictly.  A
sample is a $V$-frame clip, with $V\in[4,12]$ for mask sources and $V=4$ for
most synthetic sources.  Mask sources use one query bank per selected frame
(``each frame is a clip''), with 1,024 queries per bank.  Training uses batch
size one clip per GPU, AdamW, base learning rates $10^{-4}$ for the model and
$10^{-5}$ for the fuse encoder, 1,000 warm-up iterations, cosine decay,
gradient-norm clipping at 1.0, bf16 arithmetic, and 20,000 iterations in the
current recipe.

The sampler mixes mask, trajectory, and pseudo-static datasets using the
probabilities in the released configuration.  Representative probabilities are
0.12 (PointOdyssey), 0.10 (Kubric-4D), 0.12 (CoTracker3-Kubric), 0.11
(TAPIP3D-Kubric), 0.11 (SynthVerse), about 0.11 across the six in-house HASIM
entries, 0.066 for the later-added HASIM-static entry, and 0.0684 summed over
the seven moving-object-mask sources.  The motion branch is gated by
an explicit dynamic-dataset name list; unregistered mask datasets do not
contribute motion loss.

\paragraph{Label and prediction implementation.}
For trajectory sources, the pair label is computed online from the ground-truth
3D displacement using a 7~mm threshold.  For real mask sources, the
mask bank is sampled from 255/127/0 moving, stationary-object, and background
pixels, respectively; 127-valued stationary objects are negative, and
all-negative frames are retained.  A sixth of each bank is drawn from moving
pixels; the negatives fill the rest by taking stationary objects first, then
background, then any remaining non-moving pixel, so the hardest negatives enter
before the easy ones.  The sampler also supports a band of queries hugging the
motion boundary, which the released configuration leaves switched off.
On the five named mask entries
(DAVIS17, LASIESTA, MoCA, YouTube-VOS 2019, OVIS) the camera, depth, and
trajectory auxiliary losses
are zero-weighted, while trajectory sources use their full tracking
supervision.  The later-added HOI4D and HASIM-static entries inherit the
name-keyed defaults in the current configuration; all reported runs use these
settings.  The query-pair MLP emits one raw motion logit (the motion component of a
16-dimensional multi-task head).  The current recipe uses
$0.5\,\mathrm{BCEWithLogits}+0.5\,\mathrm{Dice}$ scaled by
$\lambda_{\mathrm{mask}}=3.0$;
identity activation is used in the head, sigmoid is applied inside the loss,
and inference thresholds the raw logit at 0.  At evaluation, a regular UV grid
at 504-pixel input resolution is decoded with downsample 4, pair logits are
aggregated by maximum, interpolated in logit space, and thresholded.

\begin{table}[h]
    \centering
    \caption{\textbf{Hyperparameters and protocol constants.}}
    \label{tab:hparams}
    \tablesetup
    \begin{tabular*}{\columnwidth}{@{\extracolsep{\fill}}llc@{}}
        \toprule
        Group & Quantity & Value \\
        \midrule
        \multirow{3}{*}{Query sampler}
          & Motion-region share $\alpha_m$     & $1/6$ (171 of 1,024) \\
          & Boundary-band share $\alpha_b$     & 0 (band disabled) \\
          & Static/background share $\alpha_h$ & $5/6$ (uniform, static first) \\
        \midrule
        \multirow{4}{*}{Objectives}
          & Dice weight $\lambda_{\mathrm{dice}}$     & 0.5 \\
          & BCE weight $\lambda_{\mathrm{bce}}$       & 0.5 \\
          & Mask loss scale $\lambda_{\mathrm{mask}}$ & 3.0 \\
          & Camera weight $\lambda_{\mathrm{cam}}$    & 0.03 \\
        \midrule
        \multirow{4}{*}{Architecture}
          & Target-memory stride                & 14~px \\
          & Target-memory tokens $N_t$          & $\lfloor H/14\rfloor\lfloor W/14\rfloor$ \\
          & Decoder blocks $B$                  & 4 \\
          & Decoder width $C$                   & 256 \\
        \midrule
        \multirow{4}{*}{Training}
          & Training queries per pair $Q$       & 1,024 \\
          & Input resolution                    & 504 px (multiples of 14) \\
          & Optimizer                           & AdamW, $10^{-4}$ ($10^{-5}$ encoder) \\
          & Schedule                            & cosine, 20k iters, 1k warm-up \\
        \midrule
        \multirow{4}{*}{Protocol}
          & Empty-mask area threshold $\tau_a$        & 20~px \\
          & Synthetic displacement threshold          & 10~mm \\
          & Ground-truth cleanup                      & $7\!\times\!7$ ellipse, 10 dilations \\
          & Baseline minimum component                & 300~px \\
        \bottomrule
    \end{tabular*}
\end{table}

\section{Evaluation Protocol and Baseline Reproduction}
\label{sec:protocol}

\paragraph{Scene-flow conversion and threshold provenance.}
The mask conversion is performed once, when the prediction PNGs are generated.
The same PNGs are then reused for all ground-truth variants, and no threshold
is reselected during evaluation.  Table~\ref{tab:flow_provenance} lists, per
method, the conversion criterion, its value, and its provenance: released
defaults for Any4D and 4RC, fully automatic criteria without a hand-set
threshold for Trace Anything and VGGT4D, and one documented substitution for
MoVieS, whose official renderer-flow value floods the frame after the
window-maximum aggregation this benchmark requires.

\begin{table}[t]
    \centering
    \caption{\textbf{Scene-flow-to-mask conversion provenance.}  ``Auto'' rows
    follow the released pipeline and expose no hand-set threshold.}
    \label{tab:flow_provenance}
    \tablesetup
    \renewcommand{\arraystretch}{1.1}%
    \begin{tabular}{@{}
        >{\raggedright\arraybackslash}p{1.6cm}
        >{\raggedright\arraybackslash}p{4.2cm}
        >{\centering\arraybackslash}p{0.9cm}
        >{\raggedright\arraybackslash}p{6.3cm}@{}}
        \toprule
        Method & Criterion & Value & Provenance and rationale \\
        \midrule
        Any4D & Scene-flow magnitude & 0.1
        & \emph{Official default.}  Released custom-data path
        (\texttt{demo\_inference.py}, \texttt{FLOW\_THRESH}) with
        \texttt{use\_scene\_flow\_mask\_refined}.  Its gates are kept: MoGe mask
        eroded $3\times$, $\pm0.5\sigma$ depth consistency, one dilation. \\
        \addlinespace[3pt]
        4RC & Max world displacement
        $\max_t\|\mathbf{x}(t)-\mathbf{x}(0)\|$ & 0.025
        & \emph{Official default.}  Released track visualizer
        (\texttt{viser\_visualizer\_track.py}, \texttt{DISP\_THRESH}). \\
        \addlinespace[3pt]
        MoVieS & Window maximum of predicted motion & 0.05
        & \emph{Substituted.}  The released \texttt{mask\_by\_motion} value of
        0.005 is calibrated for rendered splat flow and floods the frame under
        window-maximum aggregation.  We take 0.05 from an 11-value sweep over
        $[0.04,0.06]$, a plateau minimizing full-resolution PER (0.0154 at
        0.05 versus 0.1791 at 0.005). \\
        \addlinespace[3pt]
        Trace Anything & Control-point variance map,
        segment-wise log-Otsu & Auto
        & \emph{Automatic.}  Released inference protocol
        (\texttt{infer.py}); no manual parameter. \\
        \addlinespace[3pt]
        VGGT4D & QK-clustering score, window-wise adaptive multi-Otsu, then
        RefineDynMask pruning & Auto
        & \emph{Automatic.}  Released demo pipeline; no manual parameter. \\
        \midrule
        UniQuery4R \emph{(backbone)} &
        warp3d-delta magnitude & 0.005
        & \emph{Swept.}  A smooth sweep over $[0.0005,0.02]$ peaks at the same
        value under both ground-truth variants of
        Table~\ref{tab:morse5k}. \\
        \bottomrule
    \end{tabular}
\end{table}

\paragraph{Scene-flow conversion.}
For each dense-motion baseline we reduce the predicted world-coordinate
displacement between the evaluated source and target to a per-pixel magnitude
before thresholding, then apply an opening followed by a closing with the same
$7\!\times\!7$ elliptical element used to build the synthetic ground truth and
drop components below 300~px (Appendix~\ref{sec:impl}).  For Trace Anything, whose
native output is a per-pixel trajectory field rather than a single
displacement, we evaluate the trajectory at the source and target timestamps
and take the difference before applying the same reduction.

\paragraph{Pair-conditioned scoring.}
Because \method predicts one binary source mask rather than instance
identities, we union every baseline's instances before scoring rather than
adopting the official MOS-I protocol, which requires an instance association
stage.  The pairs themselves are the label-consistent moving, stationary, and
identity pairs of Section~\ref{sec:overview}.

\paragraph{Runtime measurement.}
Runtime comparisons separate the final predictor's latency
from the flow, tracking, reconstruction, segmentation, and test-time
optimization its inputs require, measured on one machine with synchronized
timing and fixed resolution and frame count.

\paragraph{Static-semantic subset.}
The subset contains annotated objects that are stationary for the evaluated
pair but belong to visually salient, commonly movable categories: parked
vehicles, resting animals, stationary people, and manipulated objects before or
after contact.  It is selected using ground-truth motion intervals and masks,
never model predictions.

\section{Union over Targets}
\label{sec:union_audit}

Equation~\ref{eq:granularity} reduces a target set to one mask with a maximum,
so a false positive produced for any single target survives into the union.
This section measures what that costs as $\mathcal{T}$ grows.  We take the
115 sequences of the five conventional MOS benchmarks that hold at least 33
frames, sample source frames at a stride of 8, and dump the logits of every
source-target pair once.  Target sets are then assembled offline, so all
columns of Table~\ref{tab:union_full} share one forward pass and differ only
in which targets are unioned.  Because the subset keeps long sequences only,
absolute values sit above Table~\ref{tab:main_mos} and the two should not be
compared.

\begin{table}[h]
    \centering
    \caption{\textbf{Union over targets}, on the five conventional MOS
    benchmarks.  Recall is monotone in $|\mathcal{T}|$ by construction;
    precision is not.}
    \label{tab:union_full}
    \tablesetup
    \begin{tabular*}{\columnwidth}{@{\extracolsep{\fill}}lccccc@{}}
        \toprule
        $|\mathcal{T}|$ & 2 & 4 & 8 & 16 & 32 \\
        \midrule
        \multicolumn{6}{@{}l}{\emph{Precision}} \\
        \quad DAVIS16-M  & 87.4 & 87.1 & 86.6 & 86.0 & 85.2 \\
        \quad DAVIS17-M  & 88.2 & 87.8 & 87.3 & 86.8 & 85.9 \\
        \quad SegTrackv2 \citep{li2013segtrackv2} & 80.8 & 79.8 & 77.8 & 75.1 & 72.6 \\
        \quad FBMS-59    & 89.8 & 89.5 & 89.2 & 89.0 & 89.0 \\
        \quad MoCA       & 87.5 & 86.0 & 83.2 & 80.3 & 76.6 \\
        \addlinespace[2pt]
        \multicolumn{6}{@{}l}{\emph{Recall}} \\
        \quad DAVIS16-M  & 93.2 & 93.4 & 93.6 & 93.8 & 94.1 \\
        \quad DAVIS17-M  & 90.2 & 91.1 & 91.8 & 92.5 & 93.3 \\
        \quad SegTrackv2 \citep{li2013segtrackv2} & 81.6 & 87.4 & 89.3 & 90.4 & 91.6 \\
        \quad FBMS-59    & 87.3 & 88.3 & 88.8 & 89.1 & 89.1 \\
        \quad MoCA       & 49.7 & 50.6 & 53.0 & 55.0 & 57.1 \\
        \addlinespace[2pt]
        \multicolumn{6}{@{}l}{\emph{$\mathcal{J}\&\mathcal{F}$}} \\
        \quad DAVIS16-M  & 83.5 & 83.3 & 83.1 & 82.7 & 82.3 \\
        \quad DAVIS17-M  & 81.5 & 81.8 & 82.0 & 82.1 & 82.0 \\
        \quad SegTrackv2 \citep{li2013segtrackv2} & 70.6 & 73.9 & 73.5 & 71.2 & 69.4 \\
        \quad FBMS-59    & 80.8 & 81.1 & 81.5 & 81.6 & 81.6 \\
        \quad MoCA       & 44.1 & 44.3 & 45.0 & 44.8 & 44.6 \\
        \bottomrule
    \end{tabular*}
\end{table}

\paragraph{False positives do not compound.}
An independent-error model would predict rapidly growing false-positive area as
more targets are unioned.  The measured union grows much more slowly because
different targets tend to fail on the same pixels.  Error therefore saturates
rather than accumulating, which makes a maximum a usable window operator.

\paragraph{The gain is concentrated in the first ring.}
The first expansion of the target set recovers useful additional motion;
subsequent targets contribute progressively more false positives, so precision
declines while recall rises.  Mean quality peaks at a modest window and then
changes little, motivating the adjacent-target protocol used in
Section~\ref{sec:datasets}.

\paragraph{Reruns and controls.}
Every baseline in Table~\ref{tab:main_mos} is rerun from released code under
our common protocol and marked $\dagger$, so the comparison is internally
consistent even where a number differs from the one published by its authors.
Baselines that consume a temporal window rather than an
isolated pair receive a \emph{matched-window} rerun in which the input window
spans the evaluated source and target frames.  The architectural controls are
the variants of Table~\ref{tab:ablation_main}, covering the low-level branch,
the auxiliary camera head, the encoder initialization, and the fusion rule.
Each is trained from the same UniQuery4R initialization and optimization
schedule as the default, so they separate the gain contributed by the
pretrained backbone from the gain contributed by the surrounding architecture.

\section{Ablation Details}

The branch, initialization, and fusion controls are in
Table~\ref{tab:ablation_main}; all variants share the data split and training
schedule.

\begin{table}[t]
    \centering
    \tablesetup
    \caption{\textbf{Per-seed spread behind Table~\ref{tab:ablation_main}.}
    Mean $\pm$ std over two seeds; the default row averages repeated runs.}
    \label{tab:ablation_spread}
    \newcommand{\pmv}[2]{\makebox[3.4em][r]{#1}\,$\pm$\,\makebox[3.0em][l]{#2}}%
    \begin{tabular*}{\columnwidth}{@{\extracolsep{\fill}}lccc@{}}
        \toprule
        Variant & wJF$\uparrow$ & PER$\downarrow$ ($\times 10^{-4}$)
        & FER$\downarrow$ (\%) \\
        \midrule
        \rowcolor{oursgray}
        Default (\method) & \pmv{77.39}{0.85} & \pmv{0.273}{0.263} & \pmv{1.45}{0.99} \\
        \midrule
        \multicolumn{4}{@{}l}{\emph{Low-level branch}} \\
        \quad ResNet-18       & \pmv{77.27}{0.70} & \pmv{0.348}{0.049} & \pmv{2.07}{0.51} \\
        \quad ResNet-34       & \pmv{77.41}{0.74} & \pmv{1.448}{1.605} & \pmv{2.64}{0.90} \\
        \quad ResNet-50       & \pmv{78.23}{0.55} & \pmv{2.428}{2.985} & \pmv{2.80}{1.46} \\
        \quad removed         & \pmv{76.38}{0.03} & \pmv{1.898}{1.212} & \pmv{2.52}{0.68} \\
        \multicolumn{4}{@{}l}{\emph{Auxiliary camera head}} \\
        \quad removed         & \pmv{76.52}{1.58} & \pmv{1.657}{1.462} & \pmv{5.47}{2.91} \\
        \multicolumn{4}{@{}l}{\emph{Encoder initialization}} \\
        \quad DA3, official   & \pmv{74.80}{1.08} & \pmv{0.250}{0.194} & \pmv{0.48}{0.10} \\
        \quad DINOv2 ViT-g/14 & \pmv{53.89}{1.12} & \pmv{225.10\phantom{0}}{38.27\phantom{0}} & \pmv{14.41}{5.71} \\
        \quad DINOv3 ViT-L/16 & \pmv{48.38}{0.96} & \pmv{213.19\phantom{0}}{27.82\phantom{0}} & \pmv{14.47}{2.17} \\
        \multicolumn{4}{@{}l}{\emph{Fusion}} \\
        \quad Ungated pooling & \pmv{76.73}{1.20} & \pmv{3.117}{1.163} & \pmv{4.43}{0.57} \\
        \quad Concat          & \pmv{76.84}{1.12} & \pmv{0.789}{0.518} & \pmv{2.60}{1.26} \\
        \bottomrule
    \end{tabular*}
\end{table}

How the ViT and convolutional pyramids are combined matters less than whether
they are combined at all.  Ungated pooling and query-level concatenation remain
within the default seed spread, whereas removing the convolutional pyramid
causes the clearest degradation.  The result supports retaining a separate
low-level branch while keeping the fusion mechanism simple.

\section{Efficiency}
\label{sec:efficiency_robustness}

\begin{table}[t]
    \centering
    \caption{\textbf{End-to-end inference cost} on one A800-80GB at bf16.
    \emph{Precompute} is joint encoding and target-memory construction,
    amortized over a four-frame window; \emph{Core} is coordinate decoding and
    mask assembly.}
    \label{tab:efficiency}
    \tablesetup
    \begin{tabular*}{\columnwidth}{@{\extracolsep{\fill}}lcccccc@{}}
        \toprule
        Model
        & Total params. (M)$\!\downarrow$
        & Precompute (ms)$\!\downarrow$
        & Core (ms)$\!\downarrow$
        & Total (ms)$\!\downarrow$
        & FPS$\!\uparrow$ & Peak mem. (GB)$\!\downarrow$ \\
        \midrule
        \rowcolor{oursgray}
        \textbf{\method} & 1,651 & 30.5 & 42.4 & 72.9 & 13.7 & 9.4 \\
        \bottomrule
    \end{tabular*}
\end{table}

\begin{table}[t]
    \centering
    \caption{\textbf{One factor varied at a time}, on one A800-80GB at bf16.
    Default setting in bold; query counts in the stride block are per frame.}
    \label{tab:efficiency_sweeps}
    \tablesetup
    \begin{tabular*}{\columnwidth}{@{\extracolsep{\fill}}lcccc@{}}
        \toprule
        Setting
        & Precompute (ms)$\!\downarrow$
        & Core (ms)$\!\downarrow$
        & Total (ms)$\!\downarrow$
        & FPS$\!\uparrow$ \\
        \midrule
        \multicolumn{5}{@{}l}{\emph{Processed shape} ($W_{\mathrm{enc}}\!=\!4$, stride 4)} \\
        \quad $280\!\times\!504$          & 26.1 & 35.4 & \phantom{0}61.5 & 16.3 \\
        \quad $504\!\times\!280$          & 26.2 & 35.4 & \phantom{0}61.7 & 16.2 \\
        \quad $\mathbf{378\!\times\!504}$ & 33.5 & 47.1 & \phantom{0}80.5 & 12.4 \\
        \quad $406\!\times\!504$          & 35.2 & 50.6 & \phantom{0}85.8 & 11.7 \\
        \quad $504\!\times\!504$          & 42.9 & 62.7 & 105.6 & \phantom{0}9.5 \\
        \midrule
        \multicolumn{5}{@{}l}{\emph{Encoding window} $W_{\mathrm{enc}}$ ($378\!\times\!504$, stride 4)} \\
        \quad $W_{\mathrm{enc}}=2$          & 46.2 & \phantom{0}26.2 & \phantom{0}72.3 & 13.8 \\
        \quad $\mathbf{W_{\mathrm{enc}}=4}$ & 33.6 & \phantom{0}47.2 & \phantom{0}80.8 & 12.4 \\
        \quad $W_{\mathrm{enc}}=8$          & 30.4 & \phantom{0}75.2 & 105.6 & \phantom{0}9.5 \\
        \quad $W_{\mathrm{enc}}=16$         & 31.5 & 122.5 & 153.9 & \phantom{0}6.5 \\
        \midrule
        \multicolumn{5}{@{}l}{\emph{Decoding stride} ($378\!\times\!504$, $W_{\mathrm{enc}}\!=\!4$)} \\
        \quad 8 \; (2,961 queries)      & 33.6 & \phantom{0}33.3 & \phantom{0}66.8 & 15.0 \\
        \quad $\mathbf{4}$ \; (11,844)  & 33.5 & \phantom{0}47.1 & \phantom{0}80.5 & 12.4 \\
        \quad 2 \; (47,628)             & 33.8 & 131.1 & 165.0 & \phantom{0}6.1 \\
        \quad 1 \; (190,512)            & 33.7 & 493.3 & 526.9 & \phantom{0}1.9 \\
        \bottomrule
    \end{tabular*}
\end{table}

Table~\ref{tab:efficiency} reports one \method inference pass on a single
A800-80GB at bf16, timed with CUDA events over 30 repeats after ten warm-up
windows.  The figure covers the single encoder, the pair memory, and the
decoder; nothing else runs at inference.  Host-side decoding and resizing add
7.6~ms per frame on the same machine and are excluded, as is metric
computation.  The full evaluation loop is slower than the model pass because it
also includes data loading, preprocessing, scoring, and window bookkeeping.

Table~\ref{tab:efficiency_sweeps} varies one factor at a time, over ten repeats
after three warm-up windows, so its default row differs slightly from the
longer protocol of Table~\ref{tab:efficiency}.  Precompute is
insensitive to decoding stride, so a single encoded pair can be interrogated at
different query densities.  Decoder cost grows with the number of queries,
while larger encoding windows increase per-pair cost because each query reads a
larger target memory; encoder amortization partly offsets this increase.  The
sweep therefore exposes a smooth accuracy--latency trade-off rather than a
change in model behavior.

The remaining controls show the same pattern: mixed-precision execution is
effectively accuracy-neutral, and image resolution affects cost through both
token count and query count.

\section{Diagnostics and Failure Cases}
\label{sec:diagnostics}

\renewcommand{\topfraction}{0.98}
\renewcommand{\textfraction}{0.01}

\begin{figure}[tb]
    \centering
    \includegraphics[width=\linewidth]{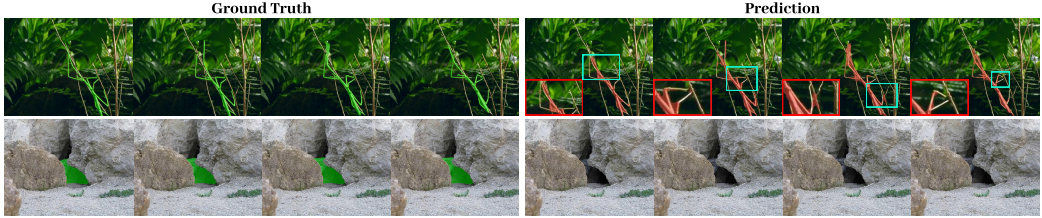}
    \caption{\textbf{Failure on small-displacement motion.}  Ground truth
    (green, left) and \method prediction (red, right) over five consecutive
    frames.  \emph{Top:} a swaying branch.  \emph{Bottom:} a small animal
    behind a rock.}
    \label{fig:limitation}
\end{figure}

Ambiguity in a single pair is a property of the formulation rather than a
failure: the pair is the smallest unit in which motion is defined, so an object
that returns to its start within the interval is read as not moving.  Synthetic
pairs whose endpoints are individually consistent with a static
scene---\emph{occluded}, \emph{ray-aligned}, and \emph{returning}---are excluded
from supervision for the same reason; pairing one source with several targets
widens the window when a longer observation is needed.  The defect this section
illustrates is the lack of spatial coupling between decoded coordinates: the
coordinate-wise decoder is most exposed on thin structures and low-texture
support, where a single query has little local evidence to draw on.

\paragraph{Small-displacement motion.}
\method does not reliably predict the mask of a region that moves by only a few
pixels.
Figure~\ref{fig:limitation} shows a branch that sways by a few pixels, which
\method recovers only in part, and a small animal that shifts behind a rock,
which it misses entirely.  Both are annotated as moving, and the region is never
activated rather than mislocalized.

\section{More Qualitative Results}
\label{sec:more_qualitative}

\begin{figure}[t]
    \centering
    \includegraphics[width=\linewidth]{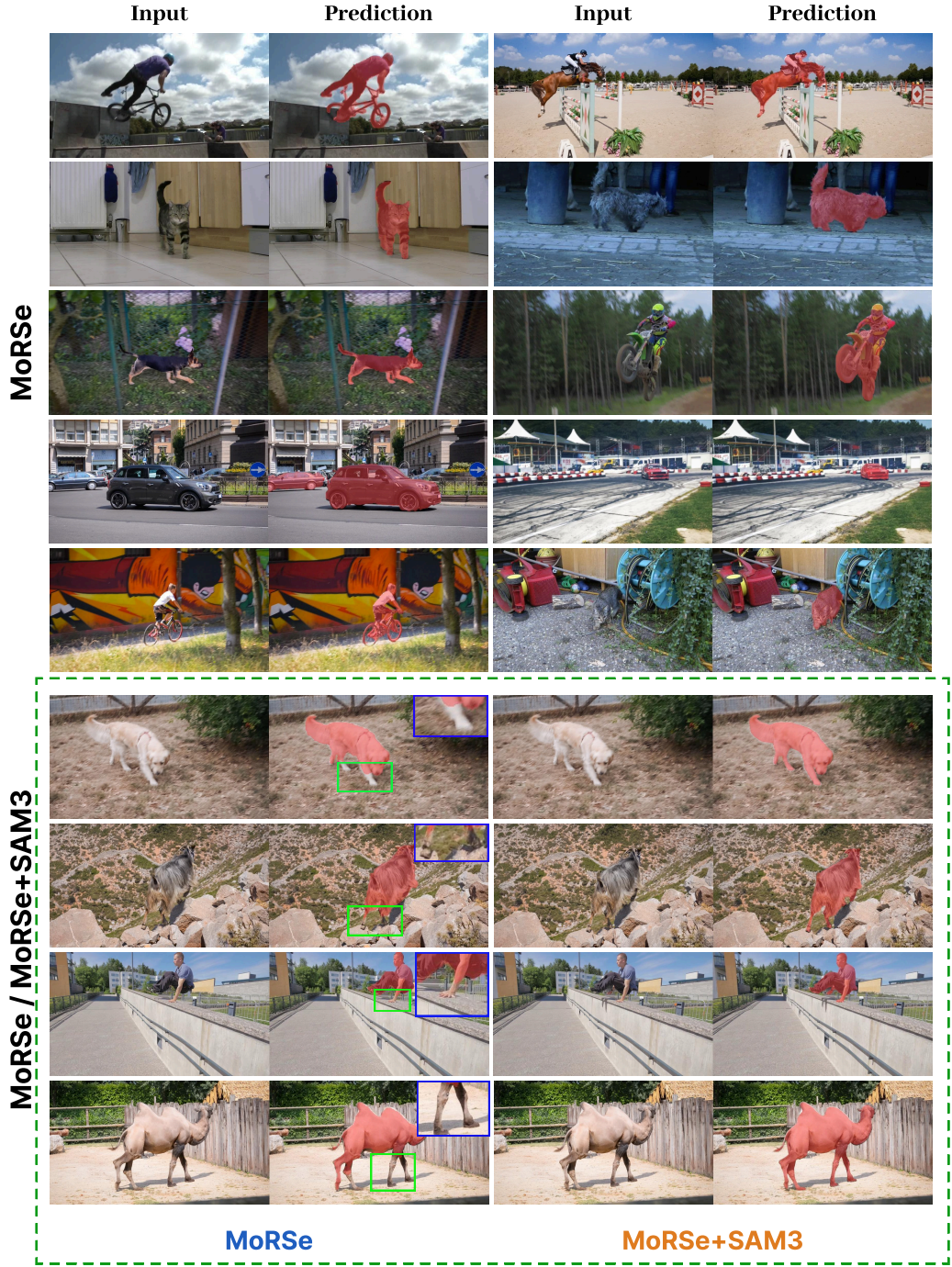}
    \caption{\textbf{Qualitative results on the conventional MOS benchmarks.}
    Columns alternate input and prediction.  \emph{Top:} \method.
    \emph{Bottom (dashed):} the same sources under \method (left) and
    \method $+$ SAM~3 (right).}
    \label{fig:more_qualitative}
\end{figure}

Figure~\ref{fig:vis_morse5k} shows \method-5K sequences in groups of one source
and three targets, where ground truth (green) and prediction (red) both follow the
target-dependent extent rather than the whole person.  On the conventional MOS
benchmarks of Table~\ref{tab:main_mos} (Figure~\ref{fig:more_qualitative}), which
label an instance as moving if any part of it does, \method recovers the moving
region---reading as an incomplete object whenever a limb or a wheel stays put---and
SAM~3 completes it to the instance the benchmark expects.

\begin{center}
    \includegraphics[width=0.9\linewidth]{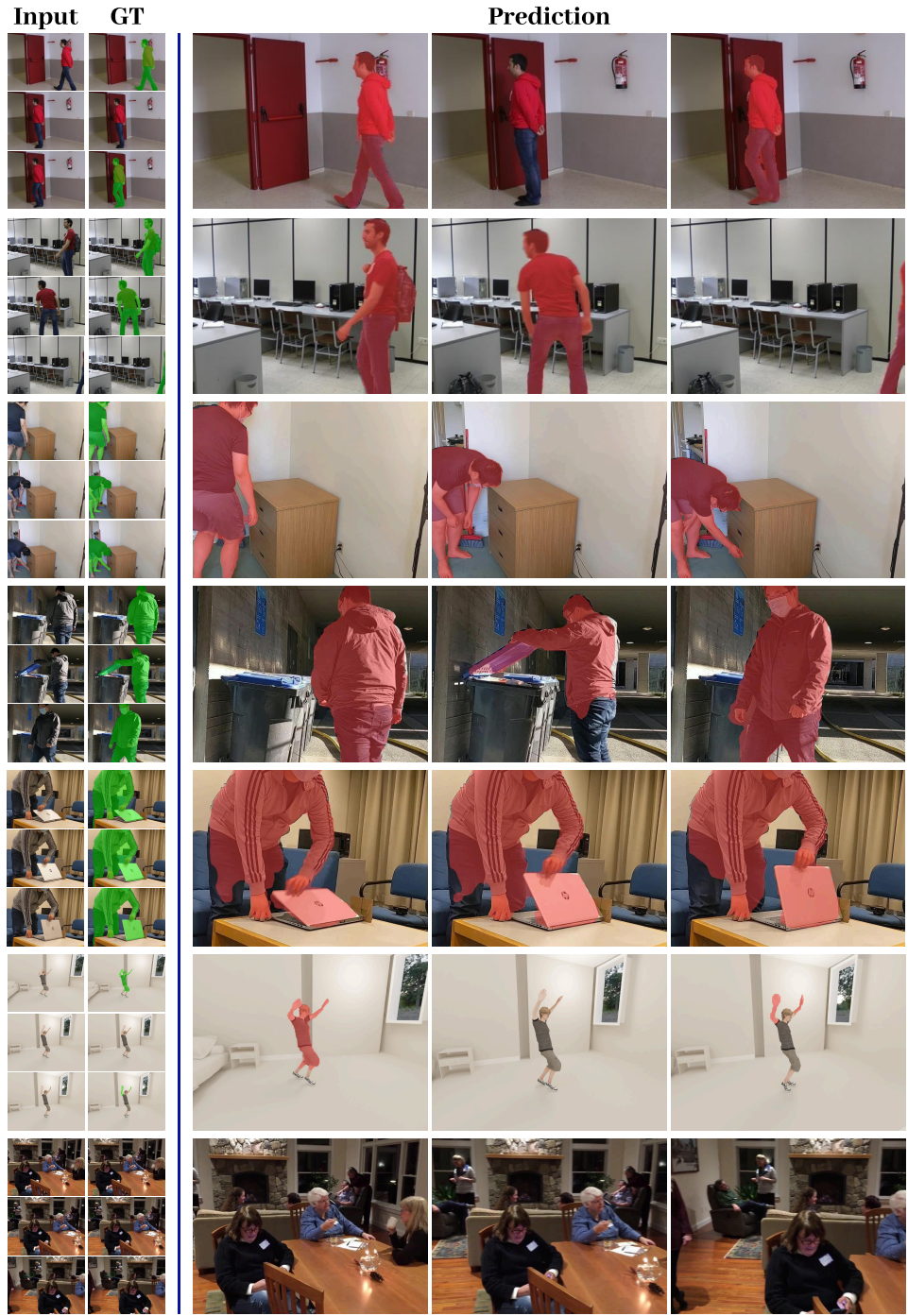}
    \captionof{figure}{\textbf{Qualitative results on \method-5K.}  \emph{Left:} source
    frames with their ground-truth moving regions (green).  \emph{Right:}
    \method predictions (red).  Within a group the source is shared and the
    target changes.}
    \label{fig:vis_morse5k}
\end{center}

\end{document}